# Recovering Lesion Parameters from Aphasic Picture Naming Error Profiles in Large Language Models

Yong Yang[1,*], Roger Newman-Norlund[2,3], Xiang Guan[1], Saeed Ahmadi[2], Regan Willis[1], Nadra Salman[5], Kalil Warren[5], Sophie Arheix-Parras[4], Srihari Nelakuditi[1], Leonardo Bonilha[6], Christopher Rorden[4], Rutvik H. Desai[4], Julius Fridriksson[2,3]

*[1]Department of Computer Science and Engineering, University of South Carolina; [2]Department of Communication Sciences and Disorders, University of South Carolina; [3]ALLT.AI, LLC; [4]Department of Psychology, University of South Carolina; [5]Linguistics Program, University of South Carolina; [6]Department of Neurology, School of Medicine, University of South Carolina*

*Corresponding author: yongy@email.sc.edu

## Abstract

Current interpretability methods for large language models (LLMs) describe internal model state but do not directly test whether the inferred state is causally sufficient to produce the observed behavior. In earlier work, we lesioned LLMs to produce error profiles in a picture naming task, which is a central task used to assess language deficits in aphasia. Specific lesions in the model produced errors that were similar to those of individual stroke survivors. Here we ask the inverse question: given an error profile, can the lesion parameters that produced it be recovered, and what does the structure of this inverse problem reveal about transformer computation? The lesions in LLaVA-Vicuna 13B were parameterized by varying layer index, modification percentage, and noise sigma across 4,840 configurations. The picture naming error profile was characterized by a clinical error taxonomy consisting of seven response types (correct, semantic, unrelated, formal, mixed, neologism, no-response). We trained a multi-task neural network to map picture naming error profiles back to perturbation parameters. The inverse problem admitted a partial solution: across 10 independently trained inverse models, both modification percentage and noise sigma were recoverable, whereas layer index was recoverable only within a neighborhood. Counterfactual validation, in which a fresh model instance was perturbed using the recovered parameters and the resulting behavior compared to the target, showed that recovered parameters reproduced target behaviors in 81.4% of cases. The dissociation between low layer recovery and high counterfactual fidelity is consistent with functional redundancy across transformer layers — an architectural property not captured by standard interpretability methods. As an out-of-distribution test of inverse-model robustness, we applied the trained model to clinical aphasia picture naming error profiles from 278 stroke survivors; recovered parameters were syndrome-discriminative — most strongly for perturbation intensity (modification percentage) — indicating that the learned representation generalizes to error-profile inputs outside its training distribution. Counterfactual validation provides a general methodological framework for LLM interpretability claims that can be applied beyond inverse mapping.

## Introduction

Interpretability methods for large language models (LLMs) (for an overview, see Bereska & Gavves, 2024) describe internal model state — through probing classifiers (Alain & Bengio, 2016; Belinkov, 2022; Hewitt & Manning, 2019), attention pattern visualization (Clark et al., 2019), causal tracing (Vig et al., 2020; Meng et al., 2022), activation patching (Wang et al., 2023), logit-lens readout of intermediate predictions (Belrose et al., 2023), and sparse autoencoders (Templeton et al., 2024) — but provide no direct test that the inferred state is causally sufficient to produce the observed behavior. They identify what correlates with behavior, not what would, if intervened upon, produce that behavior. Closing this gap requires an inverse direction: given an output, recover the lesion parameters that produced it, then verify the recovery by re-applying the predicted perturbation to a fresh model instance and checking whether the same behavior is reproduced.

Transformer architectures (Vaswani et al., 2017) permit precise control over lesions: specific layers can be modified, the fraction of affected weights specified, and perturbation intensity systematically varied. This enables dense sampling of the forward mapping from perturbations to behavior — something that has not been possible in any other neural system — and creates, to our knowledge, the first setting in which the inverse question can be tested directly in a neural network. We therefore ask: given an LLM's picture naming error profile, can the lesion parameters that produced it be recovered, and what does the structure of this inverse problem reveal about how transformers compute language?

Distinguishing forward and inverse mapping between lesion and behavior is critical. Forward studies establish what behavioral patterns arise from particular internal states by applying a perturbation and observing behavior. Inverse mapping requires the stronger claim that behavior contains sufficient information to recover the underlying perturbation, and that the recovered perturbation, when re-applied to a fresh model instance, reproduces the original behavior. This is a more stringent standard for interpretability than correlational characterization of internal state.

In earlier work (Yang et al., 2026), we established this forward direction: perturbing LLaVA-Vicuna 1.6 13B across layer, modification percentage, and noise magnitude, we matched the complete seven-category picture-naming error profile of individual stroke survivors, reproducing at least six of seven categories for 97.8% of 278 patients from a single shared perturbation space. That work showed the forward map is expressive enough to reach individual clinical profiles, but it left the inverse question open: it selected configurations by exhaustive search over the perturbation grid, never testing whether the perturbation can be recovered from behavior alone. Exhaustive search does not establish invertibility — that the behavior itself carries sufficient information to identify the perturbation that produced it. Here, we present the first inverse mapping framework for LLMs. We trained a multi-task neural network to predict perturbation parameters (layer index, modification percentage, noise sigma) from picture naming error profiles generated by a lesioned LLM. We hypothesized that it would be possible to recover the perturbation parameters given the error profiles, to the extent that each parameter combination represents a unique error profile.

## Methods

### Model and Perturbation Procedure

To enable inverse mapping, we first defined the forward perturbation procedure that specifies the parameter space to be recovered. This procedure provides a densely sampled mapping between perturbations and behavioral outputs, which is required to test whether perturbations can be recovered from behavior. We used LLaVA-Vicuna 1.6 13B (Liu et al., 2023), a multimodal model that combines a CLIP vision encoder (Radford et al., 2021) with a 40-layer Vicuna language decoder (Chiang et al., 2023). For each perturbation configuration ($L$, $M$, $\sigma$), we modified the weights of layer $L$ by applying multiplicative Gaussian noise to the weights associated with a fraction $M$ of hidden units within that layer:

$$W_{perturbed} = W_{original} \times (1 + \varepsilon), \quad \varepsilon \sim N(0, \sigma^2) \qquad (1)$$

where $W_{original}$ is the pre-trained weight matrix, $W_{perturbed}$ is the modified matrix, and $\varepsilon$ is Gaussian noise with mean zero and variance $\sigma^2$. The parameter $\sigma$ (noise sigma) controls perturbation intensity, and $M$ (modification percentage) determines the fraction of hidden units whose associated weights are modified within the targeted layer.

Perturbations were applied uniformly across all weight matrices within the selected layer, including both self-attention (query, key, value, and output projections) and MLP (gate, up, and down projections) parameters. Within each layer, hidden units were selected via random masking using a fixed seed, and all weights associated with the selected units were perturbed. This ensured that the same subset was modified for a given (L, M) pair across all noise levels. No additional random seeds were used. Gaussian noise ε was drawn independently for each perturbed weight, with the noise seed deterministically derived from the per-configuration integer identifier to maintain reproducibility.

The parameter space comprised $L$ (0–39), $M$ (0–100%, 11 values), and noise sigma $\sigma$ (1.0–2.0, 11 values), yielding 4,840 configurations. Only language decoder layers were perturbed; vision encoder parameters remained intact. Inference was conducted using greedy decoding (temperature = 0, no sampling) to ensure deterministic outputs for each configuration.

### Datasets

***Table 1.*** *Sample sizes used in analyses. All analyses use one of five analytic samples described below; each is derived from the indicated source by the specified inclusion criterion.*

| Analytic sample | Source | Inclusion criterion | Sample size |
|---|---|---|---|
| LLM training and held-out evaluation | Perturbation grid (training seed) | Stratified random 70/15/15 split of 4,840 configurations | Train 3,389 / Val 725 / Test 726 |
| Independent counterfactual validation | Independent perturbation grid (independent validation seed) | Full grid, generated under an independent random seed distinct from training/validation data | 4,840 configurations (full grid) |
| Aphasia patient (behavioral) cohort | CSTAR (Aphasia Lab, USC) | Complete 175-item PNT response record | 278 stroke surviors |
| Patient syndrome-level analysis | CSTAR subset | WAB-R diagnostic syndrome assigned; Transcortical Motor (n = 2) excluded from omnibus tests | 110 stroke survivors (108 after exclusion) |

***LLM Perturbation Dataset.*** We evaluated each perturbed configuration on the 175-item Philadelphia Naming Test (PNT; Roach et al., 1996; Schwartz et al., 2006). Each item comprises a grayscale line drawing paired with the prompt: “Name the picture using only one word. Do not use more than one word.”. Each administration includes 10 additional warmup items presented before the standard 175-item set; these warmup items are not scored and are excluded from all analyses reported here. Feature extraction aggregates responses across all 175 items per configuration, including items on which the baseline (unperturbed) model produces errors. Perturbation-induced changes on these items were retained, as they carry layer-discriminative signal. This dataset defines the forward mapping from perturbations to behavior that the inverse model seeks to invert. The dataset comprises 4,840 configurations. Data generation required approximately 2,400 GPU-hours on NVIDIA RTX 6000 Ada hardware. The dataset was split 70/15/15 into training (n = 3,389), validation (n = 725), and testing (n = 726) subsets using a fixed random seed for reproducibility.

***Independent Validation Dataset.*** To assess generalizability, we generated an independent validation dataset using a different random seed (distinct from the training-data seed). This dataset consists of all 4,840 perturbation configurations from an independently generated version of the parameter grid (40 layers × 11 modification percentages × 11 noise levels) under an independent random seed. Each configuration was evaluated on the same 175 PNT items under identical inference conditions. Full-grid (n = 4,840) counterfactual validation requires approximately 8 days on a single GPU; this cost was incurred once on a representative inverse model to provide the most stringent independent validation reasonably available. This independent generation reduces potential data leakage associated with validation on held-out subsets of a single data generation run and tests whether the inverse model captures generalizable parameter-behavior mappings rather than patterns specific to one stochastic instantiation.

***CSTAR (USC Center for the Study of Aphasia) Patient Dataset.*** Behavioral patient data were drawn from the CSTAR (Aphasia Language, Lesion, Testing, and Assessment Inventory) dataset, comprising 410 aphasia patients from the University of South Carolina. Each patient was assessed with the Western Aphasia Battery-Revised (WAB-R; Kertesz, 2007) and/or the PNT (Roach et al., 1996). The inverse model uses 7-category PNT response profiles (correct, semantic, unrelated, formal, mixed, neologism, and no-response; see Error Classification, Methods) as input; therefore, the analysis cohort included patients with complete PNT data (278 of the 410 patients). WAB-AQ scores were computed from WAB subscores (2 × [Spontaneous Speech + Comprehension + Repetition + Naming]). The CSTAR dataset provides behavioral profiles only and does not include neuroimaging data.

Diagnoses followed WAB-R criteria: Anomic (n = 24), Broca's (n = 47), Conduction (n = 15), Wernicke's (n = 5), Global (n = 5), Transcortical Motor (n = 2), and Not Aphasic (n = 12). Diagnosis labels were available for 110 patients; the remaining 168 patients lacked diagnostic classification. All participants provided informed consent under IRB-approved protocols.

## Error Classification

Model outputs were classified using a fine-tuned DeBERTa-v3-base classifier (He et al., 2021), trained on 3,160 expert-annotated (target, response) pairs derived from LLaVA-Vicuna perturbation experiments. Annotations were provided by a certified speech-language pathologist. The classifier achieved 89.2% accuracy across all eight classification categories (including correct responses) on held-out expert-annotated data. The classifier outputs eight categories corresponding to the CSTAR annotation scheme: correct, semantic, unrelated, formal, nonword, mixed, neologism, and no response. For the inverse model pipeline, the nonword and neologism categories were merged into a single neologism category, yielding seven pipeline categories. This merger reflects that both categories represent non-lexical productions, and that their distinction is less reliable for heavily perturbed LLM outputs.

## Feature Extraction

The inverse model receives a 166-dimensional behavioral profile as input for each configuration. The behavioral profile extends the underlying 7-category error profile (the foundation of standard clinical aphasia assessment) with derived statistics, response-level text features, probability moments, lexical patterns, and sentence-transformer embedding statistics computed directly from the model's actual response strings. The use of features computed from response text rather than from error labels alone distinguishes our approach from a pure inversion of the seven-category error rate vector.

For each behavioral profile, we computed features organized into six tiers, each strictly extending the previous. Tier 1 consists of seven BERT-classified error rates (correct, semantic, unrelated, formal, mixed, neologism, no response), corresponding to the categories used in clinical aphasia assessment. Tier 2 adds five statistics derived from Tier 1 through closed-form arithmetic (total error rate, linguistic and production error proportions, error pattern entropy, and an inference failure count). Tier 3 (12 features) summarizes response-level text statistics, including response length distributions and character-class proportions. Tier 4 (10 features) summarizes the moments of the per-token output probability distribution (mean, variance, skewness, kurtosis, minimum, and maximum, plus four entropy-related summary statistics); these features quantify generation-time model uncertainty and help distinguish confidently-wrong outputs (such as fluent neologisms) from low-confidence outputs (such as near-miss semantic errors). Tier 5 (8 features) measures lexical diversity, including type–token ratio, hapax-legomena rate, and perseveration indices. Tier 6 (116 features) consists of sentence-transformer embedding statistics computed using all-MiniLM-L6-v2 (Reimers & Gurevych, 2019), encoding semantic similarity between model outputs and ground-truth labels. We use the full Tier 1–Tier 6 feature set (166 features) for all main-text analyses; Supplementary Section S1 reports a controlled ablation across four feature configurations (Tier 1 only; Tiers 1–2; Tiers 1–5 without embeddings; full Tiers 1–6) under identical training protocols and twenty independent random seeds, which quantifies each tier's contribution and motivates the patient-specific simplifications described below. All features were z-score normalized using statistics fitted on the training subset.

## Inverse Model Architecture

The inverse model maps a 166-dimensional behavioral profile to perturbation parameters. The shared encoder maps the input feature vector to a 64-dimensional latent representation through three fully connected blocks (widths 256, 128, and 64). Each block consists of a linear transformation, batch normalization, ReLU activation, and dropout (p = 0.5). The three output heads operate on this shared latent representation. The layer head consists of a single layer, Linear(64, 40), producing logits for 40-way layer classification and is trained using standard cross-entropy loss. The modification-percentage head and noise-sigma heads share an identical architecture: a two-layer MLP with Linear(64, 64) followed by ReLU, and two parallel Linear(64, 1) projections that produce a predicted mean μ and log-variance log $\sigma^2$. Outputs are unbounded, and no sigmoid activation or output scaling is applied. Both regression heads are trained using a heteroscedastic Gaussian negative log-likelihood (Kendall & Gal, 2017). For the modification percentage, the loss is:

$$L_{modify} = \frac{1}{N}\sum_{i=1}^{N}\frac{1}{2}\left[log\ \sigma_i^2 + \frac{(y_i - \mu_i)^2}{\sigma_i^2}\right] \qquad (2)$$

where N is the batch size, $y_i$ is the ground-truth target value for sample i, and ($\mu_i$, log $\sigma_i^2$) are the head outputs. The total training loss is a fixed weighted sum of the three task losses: $L_{total} = L_{layer} + 0.5 \cdot L_{modify} + 0.5 \cdot L_{noise}$, where $L_{layer}$ is cross-entropy over the 40 layer classes and $L_{modify}$, $L_{noise}$ are the heteroscedastic Gaussian negative log-likelihood losses of equation (2):

$$L_{total} = 1.0 \cdot L_{layer} + 0.5 \cdot L_{modify} + 0.5 \cdot L_{noise} \qquad (3)$$

The reduced weights on the regression losses prevent their larger magnitudes from dominating the classification gradient. No correlation regularization, severity regularization, or auxiliary constraints are applied. As a result, the observed correspondence between predicted perturbation intensity and patient WAB-AQ scores arises from the data rather than from an explicit training signal. Optimization uses AdamW (Loshchilov & Hutter, 2019; learning rate $5\times10^{-4}$, weight decay $1\times10^{-2}$) with gradient clipping at 1.0 and a learning-rate scheduler that halved the rate after 20 epochs without validation-loss improvement. Training proceeds for up to 500 epochs with early stopping (patience 30); the model converged at epoch 325.

## Patient-Specific Model Application

Applying an inverse model trained on synthetic LLM perturbation data to real aphasia patient data introduces several methodological challenges. The LLM perturbation dataset and patient data differ in three respects. First, sample characteristics differ: the LLM dataset comprises 4,840 samples with known perturbation parameters, whereas the patient dataset comprises 278 samples with behavioral profiles but no corresponding ground-truth perturbations. Second, distributional structure differs: LLM perturbations sample uniformly across the parameter space, whereas patient error patterns cluster around clinically meaningful configurations. Third, feature availability differs: LLM outputs support a wide range of derived features, whereas patient data are limited to those collected in clinical assessments.

We applied a single inverse model, trained exclusively on LLM perturbation data, to the patient cohort without retraining or domain adaptation. The model architecture and weights are identical to those used for held-out LLM data. Patient PNT data provide 12 of the 166 model features (7 error rates and 5 derived statistics). The remaining 154 features (text statistics, confidence metrics, and embedding dimensions) are LLM-specific and are set to training distribution means before standardization. As a result, the model relies entirely on shared behavioral features when applied to patient data. Because the seven error rates are compositional (summing to one) and the five derived statistics are deterministic functions of these rates (with the inference-failure count fixed at zero for patients), the effective dimensionality of the patient behavioral input is approximately six independent degrees of freedom rather than twelve. The ablation reported in Supplementary Section S1 confirms that perturbation intensity prediction (modification percentage and noise sigma) is recoverable from this 12-feature subset at performance comparable to the full 166-feature representation (modification $R^2 = 0.552 \pm 0.025$ for the 12-feature model versus $0.572 \pm 0.021$ for the full model, across the 20 ablation seeds of Supplementary Table S1); perturbation location prediction is correspondingly limited.

**Syndrome-level analysis.** Of the 278 patients, 110 had an aphasia syndrome assigned by the Western Aphasia Battery–Revised (WAB-R; Kertesz, 2007), which classifies aphasia from performance on spontaneous speech, auditory comprehension, repetition, and naming: fluency and comprehension distinguish the classical syndromes, and repetition further separates conduction from other fluent presentations. Among the labeled patients, the syndromes were Broca's (n = 47), Anomic (n = 24), Conduction (n = 15), Not Aphasic (n = 12), Wernicke's (n = 5), Global (n = 5), and Transcortical Motor (n = 2). For these patients we asked whether the perturbation parameters recovered by the inverse model — which was never trained on patient data or diagnostic labels — differ systematically across the clinically defined groups. Because the syndromes index qualitatively different patterns of language breakdown, syndrome-discriminative structure in the recovered parameters would indicate that the inverse model preserves clinically meaningful distinctions rather than collapsing patients onto a single dimension. Group differences were tested with Kruskal–Wallis omnibus tests on the 108 patients in groups with n ≥ 3 (Transcortical Motor, n = 2, excluded), with effect sizes reported as $\eta^2$. This analysis is exploratory: it characterizes the recovered parameters against an external clinical taxonomy and does not assert a mechanistic correspondence between transformer layers and brain regions.

### Counterfactual Validation

Counterfactual validation (Figure 1) tests whether the inverse model captures causal parameter–behavior relationships. For each test sample, we (1) predicted parameters (L, M, σ) from the behavioral profile; (2) applied the predicted parameters to a fresh LLaVA instance not used in training; (3) evaluated the model on the same 175 PNT items; and (4) computed cosine similarity and Pearson correlation between reproduced and original error profiles across the seven classification categories. Predictions with cosine similarity > 0.85 and correlation > 0.90 were classified as high-fidelity; those with both metrics ≥ 0.70 as moderate-fidelity; the remainder as low-fidelity. The validation set consisted of all 4,840 perturbation configurations generated using an independent random seed (distinct from the training-data seed). This full-grid independent generation reduces potential data leakage and tests whether the inverse model captures generalizable parameter–behavior mappings across the entire parameter space, not a subsample of it. The counterfactual-validation procedure initially queried the model with a naming instruction ("Please name this object in one word.") that differed slightly from the instruction used to generate the perturbation dataset ("Name the picture using only one word. Do not use more than one word."). We corrected this discrepancy and confirmed that it does not affect the result: re-running a random 200-configuration subset under the generation-matched instruction reproduced the high-fidelity rate exactly (168/200 under both instructions), with marginally higher mean cosine similarity (0.926 to 0.934) and error correlation (0.903 to 0.913). The reported rate is therefore robust to, and if anything conservative under, the naming instruction.

### Multi-Seed Robustness Analysis

To assess robustness to stochastic factors in both data generation and model training, we additionally trained 10 independent inverse models on 10 separately generated datasets. Each dataset comprised an independently generated 4,840-configuration parameter grid under a distinct random seed. Each model used the same architecture and hyperparameters as the primary model; the only stochastic differences across runs were (i) the seed used for data generation, (ii) the train/validation/test split, and (iii) the random initialization and optimizer state. For each metric we report mean ± SD across the 10 models, providing a direct estimate of pipeline-wide stochastic variation.

A 10-model consensus ensemble was constructed by majority vote on layer prediction (mode across the 10 models) and mean aggregation on modification percentage and noise sigma. Ensemble performance was evaluated identically to single models and is reported alongside single-model mean ± SD (Table 3b).

Full-grid (n = 4,840) profile-level counterfactual validation was performed for both a representative single inverse model and the 10-model consensus ensemble; both are reported in Table 2. Each full-grid counterfactual run requires approximately 7 days on a single GPU. The ensemble achieved a high-fidelity rate marginally above the single model (81.9% vs. 81.4%), mirroring the small ensemble gains observed at the parameter level (Table 3b).

## Results

### Held-Out LLM Performance: Intensity is Recoverable, Location Resists

We first evaluated the inverse model on the held-out test set with known ground-truth parameters (n = 726 per model; n = 10 independently trained models). Across the 10 models, modification percentage prediction achieved $R^2 = 0.570 \pm 0.025$ and Pearson $r = 0.771 \pm 0.016$ (mean ± SD), indicating that behavioral features explain substantial variance in perturbation intensity. Noise sigma prediction showed $R^2 = 0.315 \pm 0.031$ and $r = 0.581 \pm 0.023$. Layer prediction was substantially less accurate: top-1 accuracy was 14.8% ± 1.7%, and within-5 accuracy was 77.6% ± 1.1% (Table 3). Within-5 accuracy measures whether the true layer index falls within ±5 layers of the predicted index ($|\text{predicted} - \text{true}| \leq 5$). To contextualize these values, with 40 possible layers, random prediction yields 2.5% top-1 accuracy, 27.5% within-5 accuracy, and 52.5% within-10 accuracy. The observed top-1 accuracy of 14.8% thus represents ~6× chance, and within-5 accuracy of 77.6% represents ~2.8× chance, confirming that the model captures meaningful structure despite the difficulty of layer localization.

The gap between layer-prediction accuracy (top-1 = 14.8% ± 1.7%) and counterfactual fidelity (81.4% HIGH fidelity on the full n = 4,840 independent-seed test set) reflects differences in measurement objectives. Layer top-1 accuracy evaluates exact identification among 40 layers, whereas counterfactual fidelity evaluates whether the predicted parameters reproduce the original behavioral profile within cosine similarity 0.85 and error correlation 0.90. Predictions that are inaccurate at the level of the exact layer but accurate in intensity typically yield counterfactually faithful reconstructions. The within-5 accuracy of 77.6% ± 1.1% is therefore more informative for behavioral reproduction than top-1 accuracy.

***Table 3**. Inverse model performance on held-out LLM test data, mean ± SD across 10 independently trained inverse models (n = 726 per model).*

| Target | Metric | Value | Chance |
|---|---|---|---|
| Modification % | $R^2$ | 0.570 ± 0.025 | — |
| Modification % | Pearson r | 0.771 ± 0.016 | — |
| Noise sigma | $R^2$ | 0.315 ± 0.031 | — |
| Noise sigma | Pearson r | 0.581 ± 0.023 | — |
| Layer | Top-1 accuracy | 14.8% ± 1.7% | 2.5% |
| Layer | Within-5 accuracy | 77.6% ± 1.1% | 27.5% |
| Layer | Within-10 accuracy | 88.4% ± 1.2% | 52.5% |

***Note***. *Layer prediction difficulty is consistent with functional redundancy across transformer layers, although a methodological contribution to this difficulty cannot be ruled out from this evidence alone. Chance-level baselines assume uniform random prediction over 40 layers*

To assess generalization across stochastic realizations of the data, we evaluated each of the 10 models on a full 4,840-configuration cross-seed dataset. Single-model mean performance on this independent grid was: layer top-1 = 16.3% ± 0.3%, layer within ±5 = 78.1% ± 0.6%, modification $R^2$ = 0.567 ± 0.014, noise $R^2$ = 0.330 ± 0.016 (Table 3b). The cross-seed standard deviations are notably tighter than within-distribution test variability (layer top-1 SD: 0.3% cross-seed vs. 1.7% within-distribution), indicating that the model generalizes to new stochastic realizations of the data without overfitting to any single training seed. A 10-model consensus ensemble lifted layer top-1 accuracy from 16.3% to 18.7% (a 15% relative gain), with modest gains across other metrics, demonstrating that variance across the 10 models contains independent predictive signal.

***Table 3b.*** *Cross-seed parameter-level performance on the full n = 4,840 independent-seed validation grid. Single-model column is mean ± SD across the 10 independently trained models; ensemble column is the 10-model consensus prediction.*

| Metric | Single model (n = 10, mean ± SD) | Ensemble | Δ |
|---|---|---|---|
| Layer Top-1 accuracy | 16.3% ± 0.3% | 18.7% | +2.4% |
| Layer Within-5 accuracy | 78.1% ± 0.6% | 78.5% | +0.4% |
| Modification $R^2$ | 0.567 ± 0.014 | 0.576 | +0.009 |
| Modification Pearson r | 0.771 ± 0.016 | 0.775 | +0.005 |
| Noise $R^2$ | 0.330 ± 0.016 | 0.343 | +0.013 |
| Noise Pearson r | 0.581 ± 0.023 | 0.605 | +0.024 |

*Note. Ensemble layer prediction uses majority vote (mode) across 10 models; modification and noise predictions are arithmetic means. The Δ column shows ensemble lift over the single-model mean.*

The limited layer prediction accuracy is consistent with the possibility that perturbations to different layers can produce similar behavioral profiles. Expanding the feature set from 50 features (without sentence-transformer embeddings) to 166 features (the M50 → M166 transition in the ablation study, Section S1, Table S1) yielded only modest improvements in layer prediction accuracy. This result suggests that picture naming error profiles do not uniquely specify perturbation location.

## Counterfactual Validation Confirms the Recovered Parameters Reproduce Target Behavior

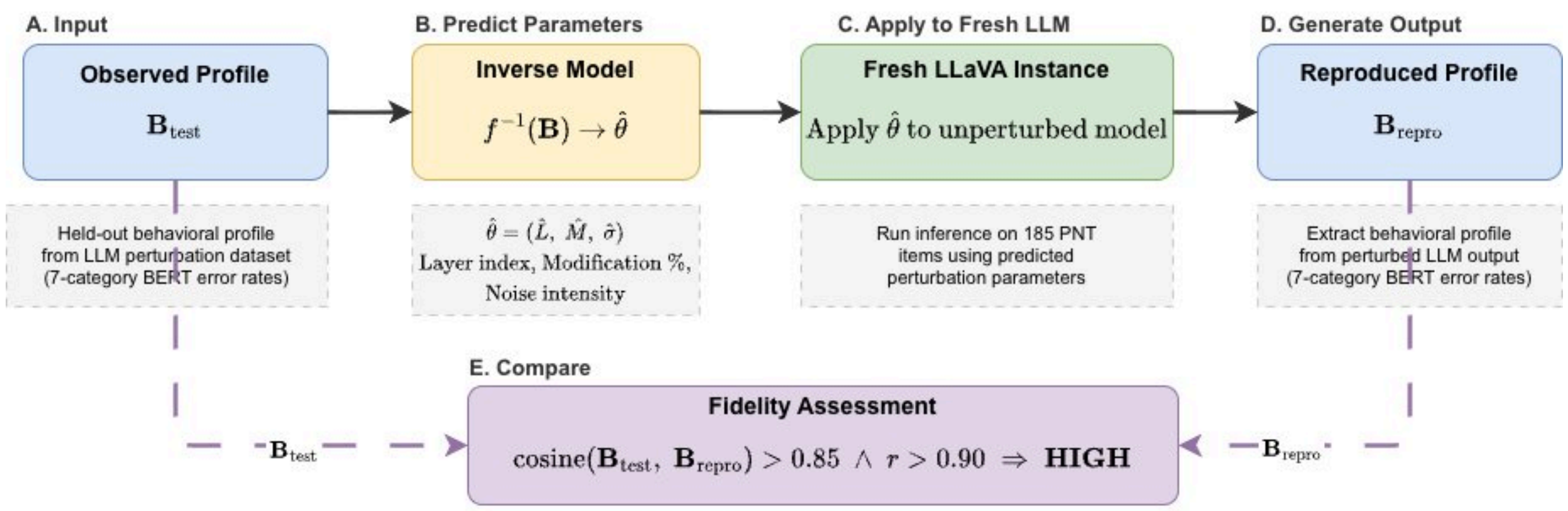

***Figure 1.*** *Counterfactual Validation Schematic. (A) An observed behavioral profile (the 7-category PNT error rates plus 5 derived statistics, comprising the 12-feature patient-applicable subset of the inverse model's 166-feature input space) serves as input. (B) The inverse model predicts perturbation parameters: layer index, modification percentage, and noise sigma. (C) Predicted parameters are applied to a fresh LLaVA-Vicuna 13B instance. (D) The reproduced behavioral profile is compared to the original using cosine similarity and error-correlation metrics. High-fidelity matching confirms that predicted parameters capture genuine parameter-behavior relationships.*

Counterfactual validation provided direct evidence that the inverse model captures genuine parameter–behavior relationships. Of all 4,840 independently generated test configurations (full grid, independent random seed), 3,938 (81.4%) achieved high-fidelity (cosine similarity > 0.85 and error correlation > 0.90) under a representative single inverse model. Mean cosine similarity was 0.9293 ± 0.1763 and mean error correlation was 0.9026 ± 0.2435 across all samples (Table 2). The 10-model consensus ensemble achieved a marginally higher high-fidelity rate of 81.9% (3,962/4,840), consistent with the small ensemble gains observed at the parameter level. These results show that predicted parameters reliably reproduce target behavioral patterns when applied to fresh model instances.

***Table 2.*** *Counterfactual validation results on the full n = 4,840 independent-seed validation grid, for a representative single inverse model and the 10-model consensus ensemble.*

| Fidelity | Criteria | Single model | | 10-model ensemble | |
|---|---|---|---|---|---|
| | | Count | % | Count | % |
| HIGH | sim > 0.85 AND corr > 0.90 | 3,938 | 81.36% | 3,962 | 81.86% |
| MODERATE | both ≥ 0.70 | 414 | 8.55% | 418 | 8.64% |
| LOW | otherwise | 488 | 10.08% | 460 | 9.50% |

***Note.*** *Single model: mean cosine similarity = 0.9293 ± 0.1763, mean error correlation = 0.9026 ± 0.2435, mean element-wise MAE = 0.0423 ± 0.0540. Ensemble: mean cosine similarity = 0.9366 ± 0.1604, mean error correlation = 0.9112 ± 0.2226, mean element-wise MAE = 0.0405 ± 0.0504. Sample size n = 4,840 (full grid) for both. The single model is one representative canonical-pipeline inverse model (data-generation seed 11003, one of the ten independently trained models); the ensemble aggregates 10 independently trained models (majority vote on layer index, mean on intensity).*

The use of an independent random seed, distinct from the training-data seed, indicates that the inverse model captures parameter–behavior relationships that generalize beyond a single stochastic instantiation.

To characterize the predictions that did not achieve high-fidelity, we examined their distribution across the parameter space. Non-high-fidelity cases were concentrated among configurations with lower modification percentages, where behavioral differences are more subtle.

## Layer Prediction vs. Counterfactual Fidelity: Functional Redundancy

We next examined whether layer-prediction correctness predicts counterfactual fidelity. On the full 4,840-sample independent validation set, predictions that correctly identified the perturbed layer (top-1 match) achieved high-fidelity reconstruction at a rate statistically indistinguishable from predictions that misidentified the layer ($\chi^2$ test, $p = 0.27$). Layer-prediction correctness did not contribute additional predictive value for counterfactual reproduction.

To quantify how much of the reproduction fidelity depends on the recovered layer versus the recovered intensity, we re-ran the counterfactual pipeline on an independent cross-seed validation set (n = 4,840) under two ablation policies and compared them against the unaltered pipeline. Randomizing the predicted layer while retaining predicted intensity reduced high-fidelity reproduction from 79.9% to 62.8%; replacing the predicted per-case intensity with the dataset-wide mean while retaining the predicted layer reduced it further to 42.4% (paired Wilcoxon signed-rank, both $p < 10^{-150}$; Supplementary Table S6.1). The intensity ablation was significantly more damaging than the layer ablation ($p \approx 10^{-118}$), confirming the intensity-over-location pattern first observed in the validation results: behavioral fidelity is governed primarily by perturbation intensity rather than by its location. This independent cross-seed replication (79.9% high-fidelity) reproduces the main-text validation result (Table 2, 81.4% high-fidelity) within seed-level variation.

Prediction errors further highlight the asymmetry between location and intensity. Non-high-fidelity cases exhibited smaller layer-prediction errors but larger modification-percentage prediction errors. Predictions that failed counterfactual reproduction were therefore more accurate in location but less accurate in intensity, an independent line of evidence consistent with the ablation result above.

These findings indicate that perturbations to different transformer layers can produce functionally equivalent outputs when perturbation intensity is matched. The behavioral signature of a perturbation therefore reflects how much the model has been disrupted, more than where the disruption was applied.

## Parameter Distributions by Syndrome

Figure 2 shows the distribution of predicted parameters across aphasia subtypes. Predicted layer varied across subtypes, with the most severe syndromes mapping to the earliest layers: Global aphasia (mean layer = 11.2, WAB-AQ = 25.2), Broca’s (12.0, 48.0), Conduction (12.7, 62.6), Not Aphasic controls (17.8, 97.3), Wernicke’s (22.0, 43.0), and Anomic (23.0, 84.2). The early-layer clustering of the most severe non-fluent syndromes is preserved, but the ordering is not strictly monotonic in severity across the milder groups.

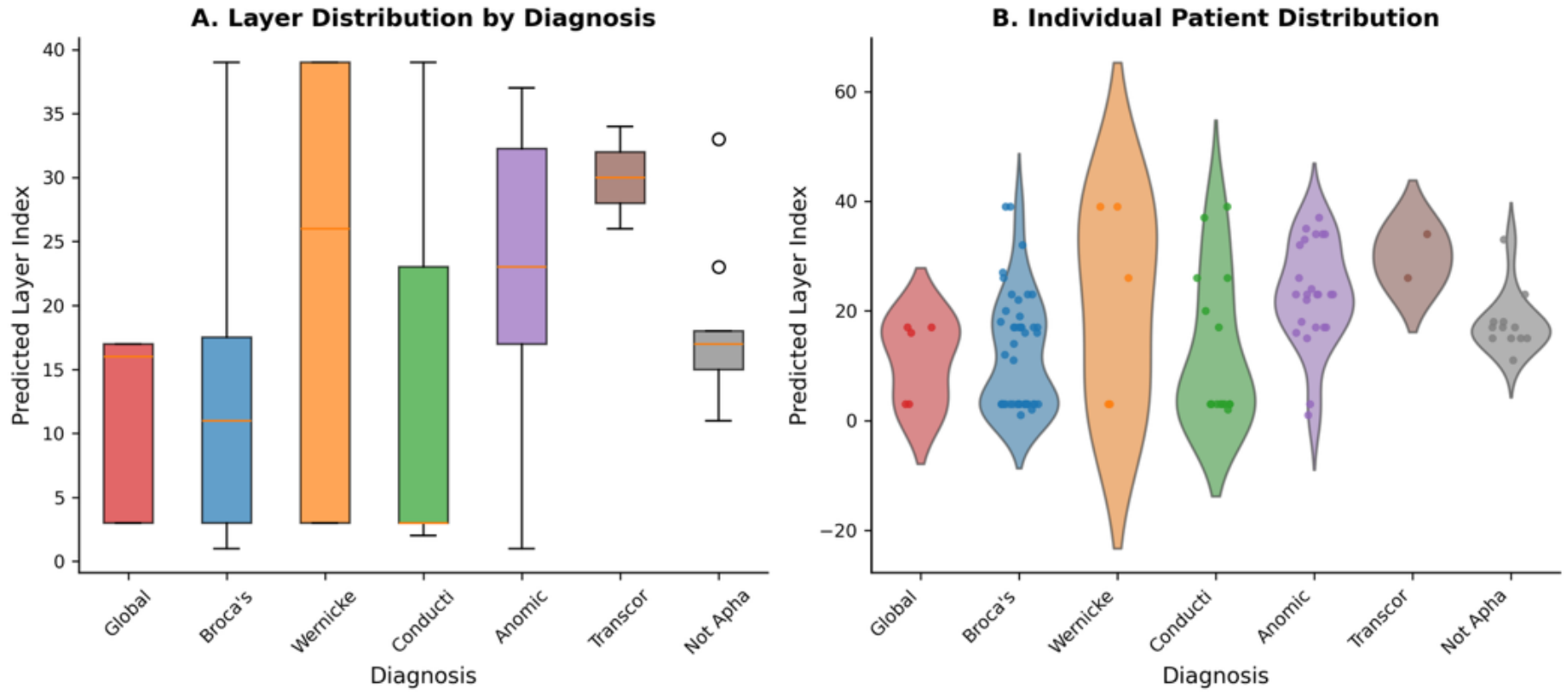

**Figure 2**. *Parameter Distributions by Diagnosis. (A) layer prediction distributions across diagnosis categories, with the most severe syndromes (Global, mean = 11.2; Broca's, 12.0) mapping to the earliest layers. (B) individual patient distribution.Sample sizes: Global (n = 5), Broca's (n = 47), Wernicke's (n = 5), Conduction (n = 15), Anomic (n = 24), Transcortical Motor (n = 2), Not Aphasic (n = 12).*

## Syndrome Signatures Confirm Discriminability

To formally test whether different aphasia syndromes occupy systematically distinct regions of the inverse-model parameter space, we compared the distributions of predicted layer index, modification percentage, and noise sigma across six syndrome groups with diagnostic labels (n = 110 of 278; see Table 1, row "Patient syndrome-level analysis"; Transcortical Motor was excluded, n = 2, below the prespecified minimum of 5). Non-parametric Kruskal–Wallis omnibus tests (Kruskal & Wallis, 1952) revealed significant syndrome-level differences for all three predicted parameters (Table 4), confirmed by 10,000-iteration permutation tests (all permutation $p < 0.001$).

**Table 4.** *Kruskal–Wallis omnibus tests for syndrome-level differences in predicted perturbation parameters (n = 108 patients across 6 syndrome groups).*

| Parameter | H | p (KW) | $\eta^2$ | Permutation p | Significant pairs (BH) |
|---|---|---|---|---|---|
| Modification % | 34.34 | < 0.001 | 0.288 | < 0.001 | 4 / 15 |
| Noise sigma | 4.63 | 0.46 | 0.00 | 0.48 | 0 / 15 |
| Layer index | 18.25 | 0.003 | 0.130 | < 0.001 | 1 / 15 |

**Note.** *$\eta^2$ = eta-squared effect size (Tomczak & Tomczak, 2014). Permutation p-values were computed with 10,000 label shuffles. Pairwise tests are Mann–Whitney U tests with Benjamini–Hochberg correction (Benjamini & Hochberg, 1995) across all 15 syndrome pairs. Transcortical Motor (n = 2) was excluded from all tests.*

Predicted modification percentage showed the largest effect ($H = 34.34$, $p < 0.001$, $\eta^2 = 0.288$; Figure 3), with mean values ascending from Not Aphasic controls (0.39) through Anomic (0.41), Conduction (0.50), Global (0.51), and the two most severe non-fluent syndromes, Wernicke's and Broca's (both 0.55). Pairwise Mann–Whitney U tests with Benjamini–Hochberg correction yielded 4 significant pairs out of 15, anchored by the Broca's vs. Not Aphasic contrast (Cliff's $\delta = 0.81$; Cliff, 1993) and the Conduction vs. Not Aphasic contrast ($\delta = 0.70$). Figure 4 summarizes these syndrome differences across all three recovered parameters, showing the group means alongside the per-patient distribution in the joint parameter space.

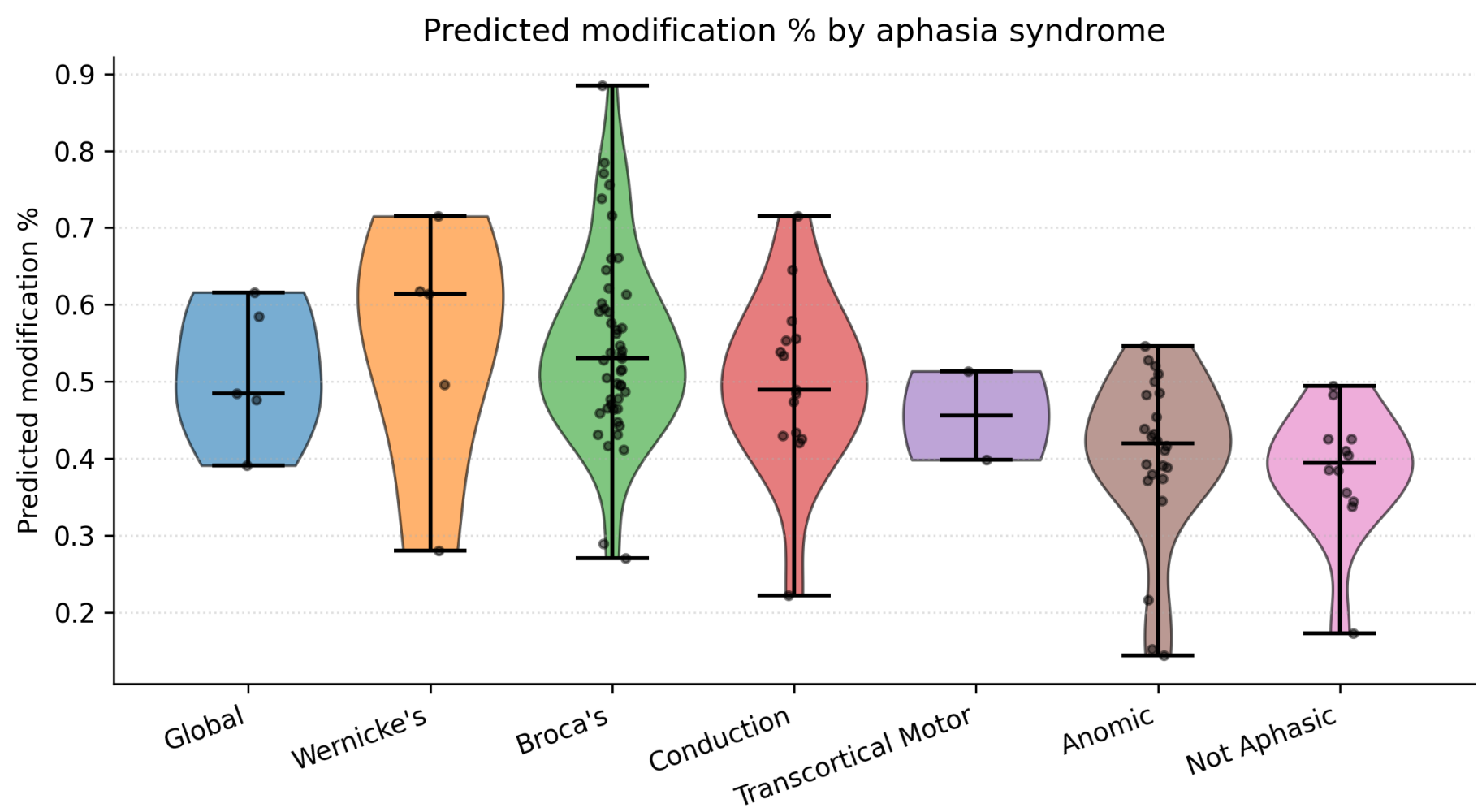

***Figure 3.*** *Predicted modification percentage by aphasia syndrome (n = 110 with diagnostic labels). horizontal bars mark the median and inner quartiles. The gradient ascends from Not Aphasic controls (mean = 0.39) through Anomic (0.41), Conduction (0.50), Global (0.51), to Wernicke's*

*and Broca's (both 0.55), tracking clinical severity. Kruskal–Wallis H = 34.34, $p < 0.001$, $\eta^2 = 0.288$. Transcortical Motor (n = 2) is shown for completeness but was excluded from statistical tests.*

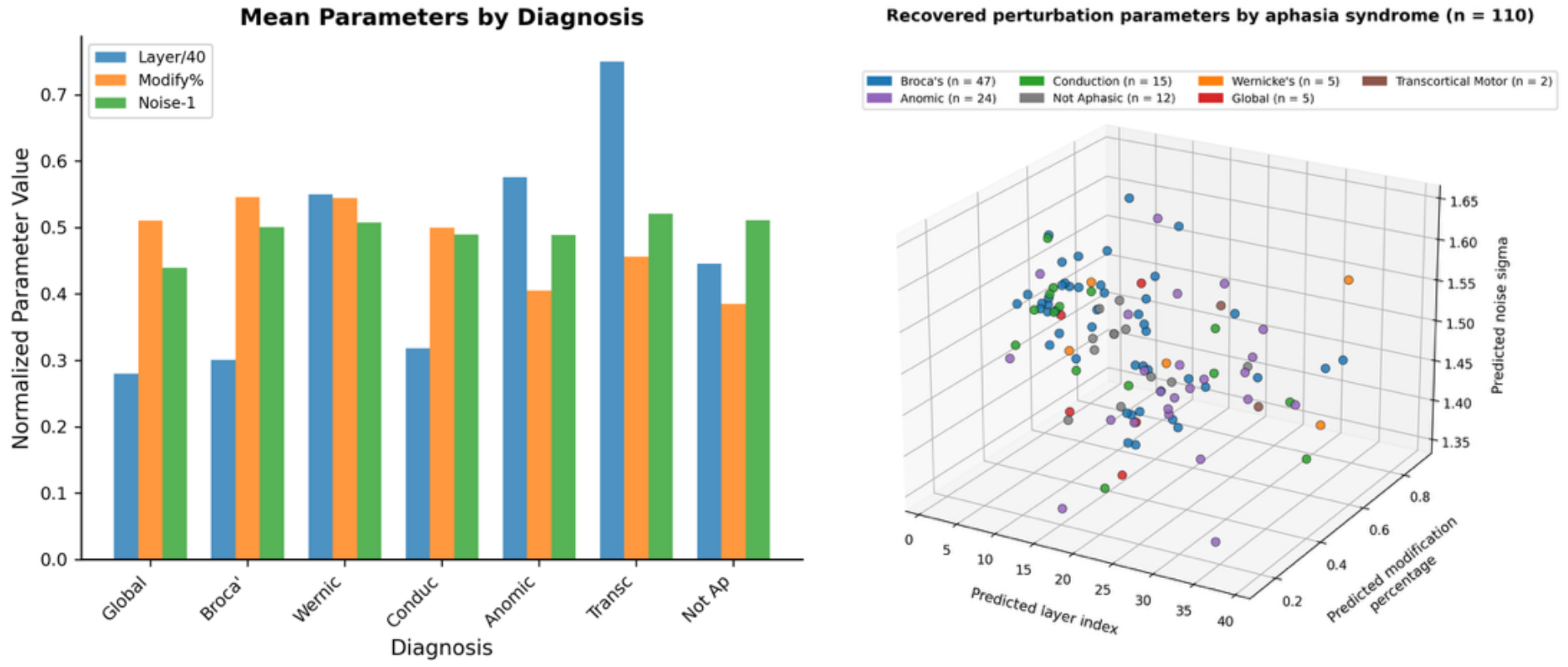

***Figure 4.*** *Recovered perturbation parameters across aphasia syndromes (n = 110 labeled patients). (Left) Mean normalized parameters — predicted layer (scaled by 1/40), modification percentage, and noise sigma minus 1 — by syndrome; the most severe syndromes (Global, Broca's) occupy the earliest predicted layers. (Right) The same recovered parameters shown per patient in the three-dimensional space of predicted layer, modification percentage, and noise sigma, colored by syndrome. Broca's patients concentrate at early layers with higher modification, whereas Anomic and Not Aphasic patients occupy later layers with lower modification. Predicted layer indices cluster at a small number of discrete values,*

Predicted layer index also discriminated syndromes (H = 18.25, $p = 0.003$, $\eta^2 = 0.130$), with the most severe non-fluent syndromes mapping to the earliest transformer layers: Global (mean = 11.2) < Broca's (12.0) < Conduction (12.7) < Not Aphasic (17.8) < Wernicke's (22.0) < Anomic (23.0). One pairwise contrast reached significance after correction (Anomic vs. Broca's, Cliff's $\delta = 0.55$). Predicted noise sigma did not differ significantly across syndromes (H = 4.63, $p = 0.46$, $\eta^2 = 0.00$; no pairwise contrast survived correction). Distributions of layer index, noise sigma, and the WAB-AQ severity gradient by syndrome are provided in the Supplementary Materials (Figure S6), with full pairwise contrast statistics in Table S3.

The ordering of effect sizes — modification % ($\eta^2 = 0.288$) > layer ($\eta^2 = 0.130$) > noise ($\eta^2 \approx 0$, n.s.) — mirrors the pattern observed in held-out LLM counterfactual validation, where perturbation intensity was the dominant determinant of behavioral fidelity and layer location was recoverable only within a neighborhood. The layer dimension also carries syndrome-discriminative information, although with a smaller effect size than modification percentage.

## Sensitivity and Robustness Analysis

Because the six-group syndrome partition includes two small groups (Global and Wernicke's, each n = 5), a single statistical test could reflect idiosyncratic features of those groups rather than a stable signal. We therefore repeated the omnibus analysis under five alternative grouping schemes designed to stress-test different potential failure modes: C1 (primary, six groups, min n = 5); C2 (merge Wernicke's and Global into a single cluster, five groups); C3 (exclude Not Aphasic controls, leaving only aphasia subtypes); C4 (strict inclusion threshold min n = 10, retaining Anomic, Broca's, Conduction, and Not Aphasic); and C5 (collapse six syndromes into three severity strata by WAB-AQ: Severe < 50, Moderate 50–85, Mild > 85). Each condition was evaluated using the same 10,000-iteration permutation test, 2,000-replicate bootstrap 95% confidence intervals on $\eta^2$, and parametric power analysis via the non-central $\chi^2$ surrogate.

**Table 5.** *Sensitivity analysis across five alternative grouping schemes. For each condition and each predicted parameter, we report the Kruskal–Wallis omnibus statistic, effect size $\eta^2$ with bootstrap 95% CI, and achieved statistical power at $\alpha = 0.05$. All 15 entries reached omnibus significance, confirming the syndrome discrimination signal is robust to grouping choices.*

| Condition | Parameter | k | n | η² [95% CI] | p (KW) | Power |
|---|---|---|---|---|---|---|
| C1 primary | Modification % | 6 | 108 | 0.288 [0.192, 0.465] | < 0.001 | >0.99 |
| C2 merge Wer/Glo | Modification % | 5 | 108 | 0.292 [0.186, 0.450] | < 0.001 | >0.99 |
| C3 excl. Not Aph. | Modification % | 5 | 96 | 0.202 [0.111, 0.390] | < 0.001 | 0.958 |
| C4 min n = 10 | Modification % | 4 | 98 | 0.326 [0.205, 0.492] | < 0.001 | >0.99 |
| C5 severity 3-fold | Modification % | 3 | 108 | 0.244 [0.132, 0.394] | < 0.001 | >0.99 |
| C1 primary | Layer index | 6 | 108 | 0.130 [0.048, 0.334] | 0.003 | 0.839 |
| C2 merge Wer/Glo | Layer index | 5 | 108 | 0.120 [0.037, 0.301] | 0.003 | 0.836 |
| C3 excl. Not Aph. | Layer index | 5 | 96 | 0.133 [0.040, 0.347] | 0.003 | 0.829 |
| C4 min n = 10 | Layer index | 4 | 98 | 0.146 [0.040, 0.340] | < 0.001 | 0.904 |
| C5 severity 3-fold | Layer index | 3 | 108 | 0.052 [0.000, 0.199] | 0.024 | 0.556 |
| C1 primary | Noise sigma | 6 | 108 | 0.00 [0.00, 0.166] | 0.46 | — |
| C2 merge Wer/Glo | Noise sigma | 5 | 108 | 0.00 [0.00, 0.128] | 0.56 | — |
| C3 excl. Not Aph. | Noise sigma | 5 | 96 | 0.00 [0.00, 0.161] | 0.45 | — |
| C4 min n = 10 | Noise sigma | 4 | 98 | 0.00 [0.00, 0.117] | 0.52 | — |
| C5 severity 3-fold | Noise sigma | 3 | 108 | 0.00 [0.00, 0.084] | 0.41 | — |

**Note.** *$k$ = number of groups used in the Kruskal–Wallis test (groups with $n$ < minimum threshold are excluded). $n$ = total analyzed sample. 95% CI were computed from 2,000 stratified bootstrap replicates. Achieved power was computed at $\alpha = 0.05$ against the observed effect size using the non-central $\chi^2$ approximation for the Kruskal–Wallis H statistic with $df = k - 1$ and non-centrality $\lambda = n \cdot \eta^2$.*

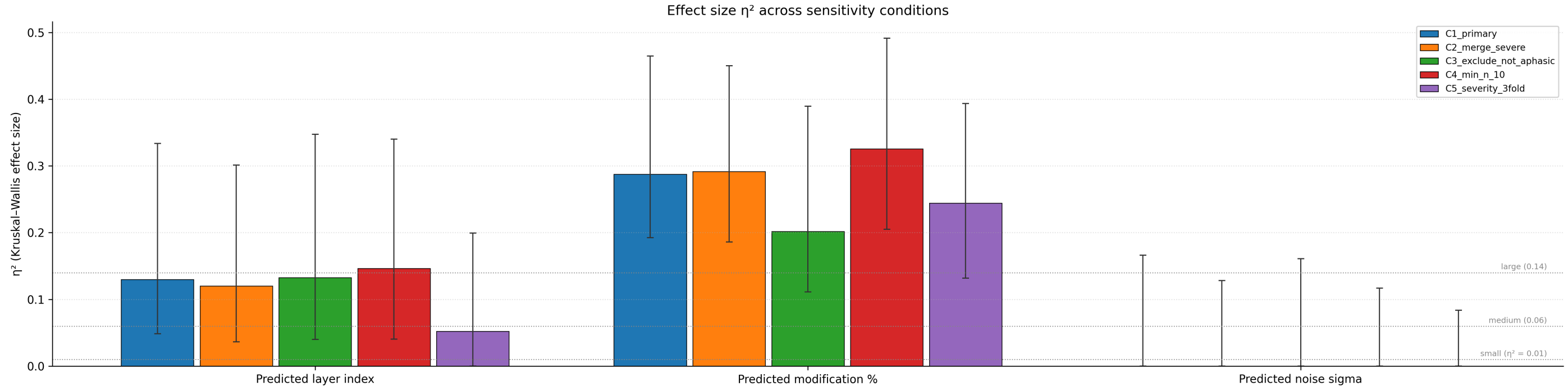

***Figure 5.*** *Effect sizes across sensitivity conditions. Grouped bars show Kruskal–Wallis $\eta^2$ for each predicted parameter (layer index, modification %, noise sigma) under the five alternative grouping schemes described in Table 5. Error bars are 2,000-replicate stratified bootstrap 95% confidence intervals. Horizontal reference lines indicate conventional small (0.01), medium (0.06), and large (0.14) effect-size thresholds. The modification-percentage and layer-index effects are significant in every condition, whereas predicted noise sigma is non-significant throughout ($\eta^2 \approx 0$); 6 of the 15 bars exceed the large-effect threshold. The C3 condition (excluding Not Aphasic controls) yields the smallest modification-percentage effect yet remains significant for both modification and layer index ($p < 0.005$), directly rejecting the hypothesis that syndrome discrimination is driven solely by the control group.*

Modification percentage and layer index yielded significant omnibus tests in every condition; predicted noise sigma was non-significant in every condition ($\eta^2 \approx 0$). The modification-percentage effect was the most stable ($\eta^2 = 0.20$–$0.33$ across conditions), followed by layer index (0.05–0.15). The C3 condition, which removes Not Aphasic controls and tests discrimination among aphasia subtypes only, produced the smallest effect sizes but retained significance for modification and layer (modification: $p < 0.001$; layer: $p = 0.003$), while noise remained non-significant ($p = 0.45$). This indicates that syndrome discrimination is not driven solely by control–aphasia contrasts and persists within the aphasia population.

The C4 condition, which retains only syndromes with $n \geq 10$ (Anomic, Broca's, Conduction, and Not Aphasic), produced the largest modification-percentage effect ($\eta^2 = 0.326$, 95% CI [0.205, 0.492]). Removing small groups increased the effect size, indicating that Wernicke's and Global groups contributed noise rather than driving the core effect. Similarly, C2 (merging Wernicke's and Global, $n = 10$) produced effects nearly identical to the primary condition ($\eta^2 = 0.292$ vs. 0.288), indicating that results are not driven by small-group artifacts.

The C5 condition collapses the six clinical syndromes into three severity strata by WAB-AQ (Severe < 50: Global, Wernicke's, Broca's, $n = 57$; Moderate 50–85: Anomic, Conduction, $n = 39$; Mild > 85: Not Aphasic, $n = 12$). If the signal were entirely driven by severity, this collapse would be expected to preserve or increase effect size. Instead, C5 yielded modestly smaller effects (modification $\eta^2 = 0.244$ vs. 0.288; layer $\eta^2 = 0.052$ vs. 0.130; noise non-significant in both), indicating that syndrome categories capture information beyond overall severity.

Power analysis confirmed that the sample size was sufficient to detect observed effects. Using the non-central $\chi^2$ surrogate for the Kruskal–Wallis statistic, the minimum detectable $\eta^2$ at 80% power and $\alpha = 0.05$ ranged from 0.089 to 0.124, below the observed modification-percentage effects. Achieved power for the modification-percentage effect exceeded 0.95 in every condition, and layer-index power exceeded 0.82 in four of the five conditions (0.56 in the three-stratum severity collapse, C5); predicted noise sigma was non-significant. Bootstrap-based subsample power curves confirmed these results: the modification-percentage signal was detectable at low subsample fractions in every condition, whereas the weaker layer-index signal required larger fractions (Figure S7).

## Discussion

Our central question, whether an LLM's picture naming error profile contains sufficient information to recover the specific perturbation that produced it, admits a more informative answer than a simple yes or no. The four findings reported here establish a partial recovery: perturbation intensity (measured by modification percentage and noise sigma) was consistently recoverable from behavior, whereas perturbation location was recoverable only within a neighborhood. Counterfactual validation on the full 4,840-configuration independent-seed grid confirmed that this partial recovery was causally sufficient to reproduce the great majority of target behavioral profiles. The dissociation between recoverable intensity and unrecoverable location is itself the most informative result, because it points to a property of transformer architecture that internal interpretability methods are unable to detect, which we examine below.

We probed the structure of this dissociation further. Across all validation cases, cases with correct layer predictions and cases with incorrect layer predictions showed indistinguishable HIGH-fidelity rates, demonstrating that counterfactual reproduction is decoupled from exact layer recovery. This decoupling identifies perturbation intensity as the primary determinant of behavioral reproduction. Overall, these patterns show that the inverse model achieves high counterfactual fidelity by recovering equivalent parameter configurations within an intensity-defined equivalence class, not by recovering the exact original layer.

Per-layer recoverability follows an asymmetric U-shaped pattern (Figure S9, Table S5): mean per-layer top-1 accuracy is highest at the vision-language embedding interface (layers 0–2: 36.1% ± 6.8 across ten training seeds, 14.4× the 2.5% chance level), falls to near the floor through the mid-stack (layers 3–35: 12.3% ± 1.6, 4.9× chance; layers 3–18 alone ≈4× chance), and rises partway again toward the output interface (layers 36–39: 19.5% ± 2.7, 7.8× chance); both the early and output-end elevations significantly exceed the mid-stack (Friedman $\chi^2 = 20.0$, $p = 4.5 \times 10^{-5}$; Wilcoxon early-vs-mid and late-vs-mid both $p = 0.002$). The early advantage is carried by layers 0 and 2 and the output-end elevation chiefly by layers 36 and 39, with a secondary elevation at layers 21–22, so the interface dominance is asymmetric, with the embedding interface the most distinctive. This pattern is consistent with functional redundancy in the middle of the transformer—many mid-stack layers produce nearly indistinguishable behavioral footprints when perturbed at matched intensity—while the embedding and output interfaces carry more distinctive computational signatures, consistent with prior evidence that information is localized to different layers across a transformer's depth (Tenney et al., 2019).

The dissociation between layer prediction accuracy and counterfactual fidelity reveals a fundamental property of transformer architecture: many layers perform functionally equivalent computations such that behavioral output is determined primarily by perturbation intensity rather than location. We tentatively label this pattern functional redundancy across transformer layers, while acknowledging that alternative explanations (such as insufficient feature richness in the inverse model) cannot be ruled out from the available evidence. The inverse model converges on this structure by identifying equivalent configurations, that is, parameter combinations that produce similar behaviors regardless of which layer is perturbed.

We tested two alternative explanations for the moderate layer-prediction accuracy that would have undermined this interpretation. First, we restricted training to PNT items on which the unperturbed baseline model produced the canonical label, under the hypothesis that baseline-incorrect items contribute noise to feature aggregation. Re-extracting features on this restricted set and retraining did not improve layer prediction accuracy. Second, we replaced the layer head's cross-entropy loss with a Gaussian-smoothed ordinal target loss, under the hypothesis that treating layer index as ordinal rather than nominal would exploit a smooth functional gradient across transformer layers. Across a range of smoothing widths, layer top-1 accuracy did not improve, and within-5 accuracy remained unchanged. These results indicate that the ceiling on exact layer prediction arises from the behavioral signature and is not removed by training-scope refinement or loss-function modification. The observed limitation therefore appears to reflect properties of the architecture rather than the training procedure, although we cannot exclude training-procedure contributions from the available evidence. This layer redundancy is not absolute. When patients are grouped by clinical syndrome, predicted layer indices show systematic differences following a severity gradient from Global aphasia to Anomic and Not Aphasic patients. Layer information is degraded at the level of individual configurations but partially preserved at the level of syndrome categories. Sensitivity analysis confirms that this layer signal is not an artifact of grouping: the effect remained significant across five alternative grouping schemes.

The counterfactual validation framework introduced here addresses a gap in the LLM interpretability literature. Existing methods, including probing classifiers, attention visualization, causal tracing, and activation patching, describe internal model state without verifying that the inferred state is causally sufficient to produce the observed behavior. A complementary approach grounds LLM interpretability in human data, using lesion–symptom mapping in stroke survivors as an external reference for which model components are functionally necessary (Fridriksson et al., 2026). Counterfactual validation closes this gap by inverting the validation direction: rather than asking whether an internal measurement correlates with behavior, it tests whether applying the inferred internal state to a fresh model instance reproduces that behavior. This procedure generalizes beyond aphasia and inverse mapping. Any interpretability claim about an LLM component can, in principle, be subjected to counterfactual evaluation by intervening on the component and testing whether the predicted behavior follows. We propose this as a methodological framework for future LLM interpretability work.

Beyond LLM-internal evaluation, we tested whether the inverse model's learned representation generalizes to error-profile inputs outside its training distribution. The training set consists of LLM outputs across all perturbation configurations; the application set consists of real-time naming responses from 278 stroke survivors, classified using the same behavior-only error taxonomy. The two distributions differ in source process (synthetic LLM perturbation versus chronic post-stroke speech production), statistical properties (uniform parameter grid versus clinically observed error-type prevalence), and item-level features (length, lexical content, response timing). If the inverse model had learned only surface statistical regularities of LLaVA outputs, application to patient data should yield noisy, uninterpretable parameter predictions. Instead, recovered parameters were structurally discriminative: they were robustly syndrome-discriminative across sensitivity analyses. The pattern observed in held-out LLM counterfactual validation — modification % > noise ≈ layer — was reproduced on the patient data, and predicted layer assignments showed a severity-aligned gradient. We make no mechanistic claim about why patient behavioral profiles produce structured signals under the LLM-trained inverse model. The result demonstrates that the learned inverse mapping is sufficiently general to recover structure from error-profile inputs it was not trained on, indicating that the inverse model's representation captures behavior-to-perturbation mapping properties not specific to the training-distribution generation process.

Several limitations warrant consideration. First, the patient sample includes few cases in two diagnostic categories (n = 5 for Wernicke's and Global), which limits within-syndrome inference. Sensitivity analysis (Table 5 and Figure 5) mitigates this concern: merging these groups (C2) or excluding them entirely (C4) produces effect sizes equal to or larger than the primary analysis, and all omnibus tests remain significant across conditions. Second, we examined a single LLM architecture, and generalization to other models remains to be established. Third, the cross-sectional design precludes investigation of recovery trajectories. Fourth, the perturbation approach modifies entire layers uniformly; finer-grained perturbations applied to individual attention heads or MLP sublayers may yield more recoverable location signals. Fifth, the counterfactual validation set (n = 4,840 full independent-seed grid), although independently generated, samples from the same parameter space as the training data, and future work should evaluate generalization to perturbation regimes outside this space. Sixth, patient inference relies on a limited set of shared features (7 error rates and 5 derived statistics), with remaining LLM-specific features set to training distribution means; this restricted overlap may constrain the achievable severity correspondence.

Future work could address several priorities. Testing additional LLM architectures would help assess the generalizability of the inverse mapping framework. Finer-grained perturbations applied to individual attention heads or MLP sublayers would test whether functional redundancy persists at sub-layer scale and may yield more recoverable location signals. Probabilistic inverse models incorporating uncertainty over equivalent parameter configurations may better capture the non-uniqueness of the inverse mapping. Expanding the behavioral evaluation beyond PNT error categories to other clinical instruments would test whether the inverse model's representation is robust to alternative behavior-only taxonomies.

In summary, we have implemented inverse mapping in a transformer and shown that picture naming error profiles contain sufficient information to recover the underlying lesion parameters: perturbation intensity is reliably recoverable, perturbation location is recoverable approximately but not precisely, and counterfactual reproduction on a fresh model instance confirms that the partial recovery is causally sufficient. The dissociation between layer recovery and counterfactual fidelity is consistent with functional redundancy across transformer layers, an architectural property that standard interpretability methods do not directly detect. As an out-of-distribution generalization test, the LLM-trained inverse model produced syndrome-discriminative outputs when applied to clinical aphasia behavioral data, indicating that the learned inverse mapping captures behavior-to-perturbation properties that are not specific to the training distribution. The methodology developed here — predicting lesion parameters from a picture naming error profile, then counterfactually verifying the prediction by re-applying the predicted perturbation to a fresh model instance — provides a general framework for interpretability claims in LLMs, applicable beyond inverse mapping to any internal-state hypothesis that can be operationalized as an intervention.

A central limitation is that our analysis evaluates a single model, LLaVA-Vicuna 1.6 13B. The dissociation we report — recoverable perturbation intensity alongside neighborhood-level layer recoverability — may reflect architectural properties specific to this model rather than a general feature of large language models. Transformer families differ in depth, width, attention structure, training data, and the degree of functional redundancy across layers, any of which could alter how much of a lesion is recoverable from a picture naming error profile. The present results should therefore be read as an existence proof that inverse mapping is feasible and informative for one representative multimodal model, not as a claim about language models in general. Establishing whether the same inverse structure holds across architectures will require replicating this procedure on models of different scale and design; this is a direction we are pursuing in subsequent work within the broader framework of which this study forms one component.

## References

Alain, G., & Bengio, Y. (2016). Understanding intermediate layers using linear classifier probes. arXiv:1610.01644

Belinkov, Y. (2022). Probing classifiers: Promises, shortcomings, and advances. Computational Linguistics, 48(1), 207–219. https://doi.org/10.1162/coli_a_00422

Belrose, N., Ostrovsky, I., McKinney, L., Furman, Z., Smith, L., Halawi, D., Biderman, S., & Steinhardt, J. (2023). Eliciting latent predictions from transformers with the tuned lens. arXiv:2303.08112

Benjamini, Y., & Hochberg, Y. (1995). Controlling the false discovery rate: A practical and powerful approach to multiple testing. Journal of the Royal Statistical Society Series B, 57(1), 289–300. https://doi.org/10.1111/j.2517-6161.1995.tb02031.x

Bereska, L., & Gavves, E. (2024). Mechanistic interpretability for AI safety — A review. Transactions on Machine Learning Research. arXiv:2404.14082

Chiang, W.-L., Li, Z., Lin, Z., Sheng, Y., Wu, Z., Zhang, H., Zheng, L., Zhuang, S., Zhuang, Y., Gonzalez, J. E., Stoica, I., & Xing, E. P. (2023). Vicuna: An open-source chatbot impressing GPT-4 with 90%* ChatGPT quality. https://lmsys.org/blog/2023-03-30-vicuna/

Clark, K., Khandelwal, U., Levy, O., & Manning, C. D. (2019). What does BERT look at? An analysis of BERT's attention. Proceedings of the 2019 ACL Workshop BlackboxNLP: Analyzing and Interpreting Neural Networks for NLP, 276–286. https://doi.org/10.18653/v1/W19-4828

Cliff, N. (1993). Dominance statistics: Ordinal analyses to answer ordinal questions. Psychological Bulletin, 114(3), 494–509.

Fridriksson, J., Newman-Norlund, R., Ahmadi, S., Willis, R., Salman, N., Warren, K., Guan, X., Yang, Y., Nelakuditi, S., Desai, R. H., Bonilha, L., Charney, J., & Rorden, C. (2026). Stroke lesions as a Rosetta Stone for language model interpretability. arXiv:2602.04074. https://doi.org/10.48550/arXiv.2602.04074

Gibson, M., Newman-Norlund, R., Bonilha, L., Fridriksson, J., Hickok, G., Hillis, A. E., den Ouden, D.-B., & Rorden, C. (2024). The Aphasia Recovery Cohort, an open-source chronic stroke repository. Scientific Data, 11(1), 981. https://doi.org/10.1038/s41597-024-03819-7

He, P., Gao, J., & Chen, W. (2021). DeBERTaV3: Improving DeBERTa using ELECTRA-style pre-training with gradient-disentangled embedding sharing. arXiv:2111.09543. https://doi.org/10.48550/arXiv.2111.09543

Hewitt, J., & Manning, C. D. (2019). A structural probe for finding syntax in word representations. Proceedings of NAACL-HLT 2019, 4129–4138. https://doi.org/10.18653/v1/N19-1419

Kendall, A., & Gal, Y. (2017). What uncertainties do we need in Bayesian deep learning for computer vision? Advances in Neural Information Processing Systems, 30. arXiv:1703.04977

Kertesz, A. (2007). Western Aphasia Battery–Revised. Pearson.

Kruskal, W. H., & Wallis, W. A. (1952). Use of ranks in one-criterion variance analysis. Journal of the American Statistical Association, 47(260), 583–621. https://doi.org/10.1080/01621459.1952.10483441

Liu, H., Li, C., Wu, Q., & Lee, Y. J. (2023). Visual instruction tuning. Advances in Neural Information Processing Systems, 36. arXiv:2304.08485

Loshchilov, I., & Hutter, F. (2019). Decoupled weight decay regularization. International Conference on Learning Representations. arXiv:1711.05101

Meng, K., Bau, D., Andonian, A., & Belinkov, Y. (2022). Locating and editing factual associations in GPT. Advances in Neural Information Processing Systems, 35, 17359–17372.

Radford, A., Kim, J. W., Hallacy, C., Ramesh, A., Goh, G., Agarwal, S., Sastry, G., Askell, A., Mishkin, P., Clark, J., Krueger, G., & Sutskever, I. (2021). Learning transferable visual models from natural language supervision. Proceedings of the 38th International Conference on Machine Learning, PMLR 139, 8748–8763.

Reimers, N., & Gurevych, I. (2019). Sentence-BERT: Sentence embeddings using Siamese BERT-networks. Proceedings of the 2019 Conference on Empirical Methods in Natural Language Processing (EMNLP-IJCNLP), 3982–3992. https://doi.org/10.18653/v1/D19-1410

Roach, A., Schwartz, M. F., Martin, N., Grewal, R. S., & Brecher, A. (1996). The Philadelphia Naming Test: Scoring and rationale. Clinical Aphasiology, 24, 121–133.

Schwartz, M. F., Dell, G. S., Martin, N., Gahl, S., & Sobel, P. (2006). A case-series test of the interactive two-step model of lexical access: Evidence from picture naming. Journal of Memory and Language, 54(2), 228–264. https://doi.org/10.1016/j.jml.2005.10.001

Templeton, A., Conerly, T., Marcus, J., Lindsey, J., Bricken, T., Chen, B., Pearce, A., Citro, C., Ameisen, E., Jones, A., Cunningham, H., Turner, N. L., McDougall, C., MacDiarmid, M., Tamkin, A., Durmus, E., Hume, T., Mosconi, F., Freeman, C. D., Sumers, T. R., Rees, E., Batson, J., Jermyn, A., Carter, S., Olah, C., & Henighan, T. (2024). Scaling monosemanticity: Extracting interpretable features from Claude 3 Sonnet. Transformer Circuits Thread. https://transformer-circuits.pub/2024/scaling-monosemanticity/

Tenney, I., Das, D., & Pavlick, E. (2019). BERT rediscovers the classical NLP pipeline. Proceedings of the 57th Annual Meeting of the Association for Computational Linguistics, 4593–4601. https://doi.org/10.18653/v1/P19-1452

Tomczak, M., & Tomczak, E. (2014). The need to report effect size estimates revisited: An overview of some recommended measures of effect size. Trends in Sport Sciences, 21(1), 19–25.

Vaswani, A., Shazeer, N., Parmar, N., Uszkoreit, J., Jones, L., Gomez, A. N., Kaiser, Ł., & Polosukhin, I. (2017). Attention is all you need. Advances in Neural Information Processing Systems, 30. arXiv:1706.03762

Vig, J., Gehrmann, S., Belinkov, Y., Qian, S., Nevo, D., Singer, Y., & Shieber, S. (2020). Investigating gender bias in language models using causal mediation analysis. Advances in Neural Information Processing Systems, 33. arXiv:2004.12265

Wang, K., Variengien, A., Conmy, A., Shlegeris, B., & Steinhardt, J. (2023). Interpretability in the wild: A circuit for indirect object identification in GPT-2 small. International Conference on Learning Representations. arXiv:2211.00593

Yang, Y., Guan, X., Arheix-Parras, S., Ahmadi, S., Newman-Norlund, R., Bonilha, L., Rorden, C., Fridriksson, J., Desai, R. H., & Nelakuditi, S. (2026). Lesioned multimodal language models reproduce aphasic picture-naming patterns. arXiv:2607.11621. https://doi.org/10.48550/arXiv.2607.11621

## Acknowledgements

We thank the patients who participated in this research and the CSTAR team for behavioral data collection. This work was supported by NIH/NIDCD P50 DC014664 (JF) and DC017162 (RHD). Computational resources were provided by the University of South Carolina Research Computing Center.

## Competing Interests

The authors declare no competing interests.

## Use of Generative AI

Generative AI tools (large language model assistants) were used solely as productivity aids in preparing this work: to assist with drafting and language editing of the manuscript, to help develop and debug the analysis code, and to help prepare figures, tables, and the accompanying code-release documentation. All research ideas, study design, algorithms, analytic logic, interpretation, and verification are the authors' own. No AI-generated data or results were included in the paper; every reported result was produced by the authors' own executed and manually verified computations. The authors reviewed and edited all AI-assisted output and take full responsibility for the content of this work.

## Data Availability

The LLM perturbation dataset—including the raw model outputs, the derived feature tables, and the trained inverse models—is available from the corresponding author upon reasonable request and will be archived in a public repository with a permanent DOI upon publication. The CSTAR patient data are not publicly available because they contain sensitive clinical information; de-identified data may be made available to qualified researchers upon reasonable request, subject to institutional review board approval and a data use agreement.

## Code Availability

The code used to generate the perturbation dataset, extract the behavioral and embedding features, train the multi-task inverse model, and perform the counterfactual and patient-application analyses is available from the corresponding author upon reasonable request. Upon peer-reviewed publication, the complete analysis pipeline will be released in a public repository and archived with a permanent DOI.

## Supplementary Information

Supplementary Information is provided as a separate document and consists of six sections: Section S1, Feature Pipeline Ablation (the controlled feature-tier ablation referenced in Methods); Section S2, Diagnosis-Level Prediction Diagnostics; Section S3, Syndrome Signature Analysis; Section S4, Sensitivity and Robustness Analysis; Section S5, Multi-Seed Robustness (per-layer accuracy bar chart with error bars across 10 independent training seeds; per-seed metrics); and Section S6, Counterfactual Baseline Ablation (the layer- and intensity-ablation baseline for the counterfactual validation). All Supplementary Figures and Tables cited in the main text are contained therein.

# Supplementary Information for: Recovering Lesion Parameters from Aphasic Picture Naming Error Profiles in Large Language Models

## Section S1. Feature Pipeline Ablation

This section reports a controlled ablation study quantifying the contribution of each feature tier to inverse-model performance. The ablation motivates two methodological choices in the main paper: (i) the use of the full 166-feature representation for all main-text analyses, and (ii) the recognition that perturbation-intensity recovery, which the main paper validates by counterfactual reproduction, depends predominantly on the lower feature tiers (text statistics, probability moments, lexical patterns) rather than the high-dimensional sentence-transformer embeddings.

### S1.1 Methods

Four nested feature configurations were defined, each strictly extending the previous: M7 (Tier 1 only: seven classifier-labeled error rates); M12 (Tiers 1–2: error rates plus five derived statistics computed from the error rates by closed-form arithmetic); M50 (Tiers 1–5: error rates, derived statistics, output text statistics, prediction probability moments, and lexical-diversity features; 50 features total, no embeddings); M166 (Tiers 1–6: full pipeline including sentence-transformer embedding features; 166 features total).

All four models share the identical inverse model architecture, training protocol, and hyperparameters used in the main paper: a three-layer MLP encoder (widths 256, 128, 64) with batch normalization, ReLU activation, and dropout $p = 0.5$; AdamW optimization (learning rate $5\times10^{-4}$, weight decay $1\times10^{-2}$); batch size 64; cross-entropy loss for the layer head and heteroscedastic Gaussian negative log-likelihood for modification-percentage and noise-sigma heads. Only the input feature dimension differs across models. Early-stopping patience was set to 100 epochs (maximum 1000 epochs). Each configuration was trained from twenty independent random seeds (42, 123, 456, 789, 1024, 2048, 4096, 8192, 16384, 271, 314, 577, 1414, 1729, 2718, 3141, 6022, 8765, 9001, 9876). The same 4,840 perturbation configurations are used across all eighty runs.

Statistical comparison between consecutive model pairs used the paired-sample t-test on each metric, with pairing by training seed. Cohen's $d_6$ is reported as the effect size for paired data, computed as the mean of the per-seed differences divided by their standard deviation. P-values were corrected using the Benjamini–Hochberg false discovery rate procedure across all forty-eight pairwise comparisons (three model pairs × sixteen metrics).

### S1.2 Results

Performance across the four configurations on key intensity- and location-prediction metrics is summarized in Table S1 and visualized in Figure S1, with the pairwise paired-sample statistics in Table S2. Three transitions are visible in the data, and each has a different character.

M7 → M12 (adding Tier 2 derived statistics). Adding five statistics computed from Tier 1 by closed-form arithmetic produced small but mixed effects. Layer top-1 accuracy was statistically indistinguishable ($\Delta = -0.0001$, $p = 0.98$, $d_6 = -0.01$), and modification $R^2$ increased modestly but did not reach significance ($\Delta = +0.0075$, $p = 0.12$, $d_6 = 0.37$). Tier 2 features contain information but, because they are deterministic functions of Tier 1, are largely redundant for a neural network with sufficient capacity to learn equivalent transformations directly from Tier 1.

M12 → M50 (adding Tiers 3–5). This was the dominant transition. Every metric improved with effect sizes ranging from $d_6 = 1.59$ (noise $R^2$) to $d_6 = 14.87$ (layer top-5), all BH-corrected $p < 0.001$. Layer top-1 accuracy increased from 4.92% to 13.71% ($\Delta = +8.79$ percentage points). Intensity metrics also improved: modification $R^2$ rose from 0.552 to 0.596 and noise $R^2$ from 0.241 to 0.345. These three tiers contribute the bulk of the inverse model's predictive signal.

M50 → M166 (adding Tier 6 sentence-transformer embeddings). This transition revealed a statistically significant trade-off rather than a uniform improvement. Layer- and region-prediction metrics improved (layer top-1 +0.96 pp, region top-1 +4.04 pp, both BH-corrected $p < 0.05$). Intensity-prediction metrics moved in the opposite direction: modification $R^2$ decreased from 0.596 to 0.572 ($\Delta = -0.024$, $p < 0.001$, $d_6 = -1.65$), noise $R^2$ decreased from 0.345 to 0.326 ($\Delta = -0.019$, $p < 0.001$, $d_6 = -1.31$). The full 166-feature representation is retained for main-text analyses for continuity with prior LLM-perturbation pipelines and to preserve the layer- and region-prediction improvements; the intensity trade-off is small in absolute terms ($R^2$ differences of order 0.02) and does not affect the qualitative conclusions of the counterfactual validation reported in the main paper.

*Table S1. Inverse model performance by feature configuration (mean ± SD across 20 random seeds). All configurations were trained for up to 1000 epochs with early-stopping patience 100; mean training time per run was approximately 100 s on an NVIDIA RTX 6000 Ada GPU. Parenthetical values in column headings indicate the number of input features.*

| Metric | M7 (7 feat.) | M12 (12 feat.) | M50 (50 feat.) | M166 (166 feat.) |
|---|---|---|---|---|
| Layer top-1 accuracy | 4.92% ± 0.69 | 4.92% ± 0.71 | 13.71% ± 1.19 | 14.66% ± 0.99 |
| Layer within-5 accuracy | 35.05% ± 2.20 | 37.49% ± 2.42 | 73.98% ± 1.99 | 77.71% ± 1.60 |
| Layer MAE (layers) | 12.64 ± 0.47 | 12.09 ± 0.58 | 5.04 ± 0.45 | 4.62 ± 0.45 |
| Layer Pearson r | 0.314 ± 0.030 | 0.310 ± 0.027 | 0.749 ± 0.044 | 0.767 ± 0.041 |
| Region top-1 accuracy | 31.10% ± 1.57 | 31.18% ± 1.66 | 61.15% ± 1.62 | 65.19% ± 1.63 |
| Modification $R^2$ | 0.545 ± 0.013 | 0.552 ± 0.025 | 0.596 ± 0.016 | 0.572 ± 0.021 |
| Modification Pearson r | 0.740 ± 0.008 | 0.747 ± 0.011 | 0.777 ± 0.011 | 0.776 ± 0.012 |
| Noise $R^2$ | 0.246 ± 0.035 | 0.241 ± 0.064 | 0.345 ± 0.028 | 0.326 ± 0.030 |
| Noise Pearson r | 0.501 ± 0.033 | 0.509 ± 0.034 | 0.590 ± 0.025 | 0.583 ± 0.027 |

*Table S2. Paired-sample pairwise comparison statistics (twenty seeds per pair, Benjamini–Hochberg-corrected across forty-eight comparisons). Δ mean = mean(model B) − mean(model A) across twenty seeds. Pairing is by training seed.*

| **Comparison** | **Metric** | **Δ mean** | **t** | **p (BH)** | **Cohen's $d_6$** |
|---|---|---|---|---|---|
| M12 vs M7 | Layer top-1 | −0.0001 | −0.03 | 0.98 | −0.01 |
| | Layer within-5 | +0.0244 | 6.14 | < 0.001 | +1.37 |
| | Layer Pearson r | −0.0040 | −1.02 | 0.36 | −0.23 |
| | Modification $R^2$ | +0.0075 | 1.64 | 0.13 | +0.37 |
| | Noise $R^2$ | −0.0045 | −0.32 | 0.80 | −0.07 |
| M50 vs M12 | Layer top-1 | +0.0879 | 31.12 | < 0.001 | +6.96 |
| | Layer within-5 | +0.3649 | 62.09 | < 0.001 | +13.89 |
| | Layer Pearson r | +0.4395 | 42.24 | < 0.001 | +9.44 |
| | Modification $R^2$ | +0.0435 | 7.42 | < 0.001 | +1.66 |
| | Noise $R^2$ | +0.1033 | 7.12 | < 0.001 | +1.59 |
| M166 vs M50 | Layer top-1 | +0.0096 | 2.72 | 0.017 | +0.61 |
| | Layer within-5 | +0.0373 | 8.99 | < 0.001 | +2.01 |
| | Layer Pearson r | +0.0175 | 2.45 | 0.029 | +0.55 |
| | Region top-1 | +0.0404 | 11.83 | < 0.001 | +2.65 |
| | Modification $R^2$ | −0.0239 | −7.38 | < 0.001 | −1.65 |
| | Noise $R^2$ | −0.0189 | −5.88 | < 0.001 | −1.31 |

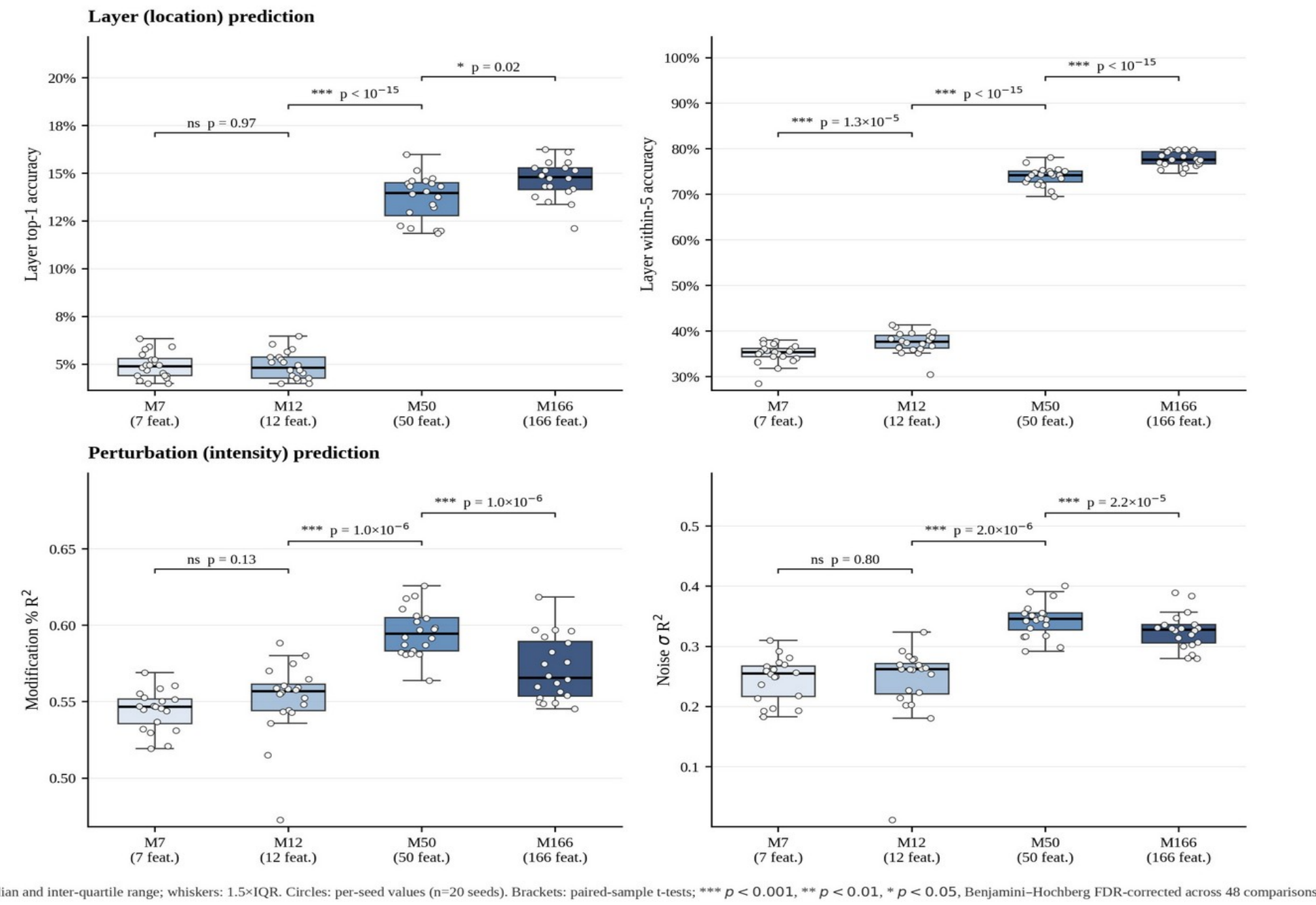

***Figure S1.*** *Inverse-model performance across four feature configurations on key prediction targets. Top row: layer (location) prediction; bottom row: perturbation (intensity) prediction. Each panel shows the distribution of metric values across twenty independent training seeds for each model (M7, M12, M50, M166). Boxes show median and inter-quartile range; whiskers extend to 1.5×IQR. Circles mark per-seed values with horizontal jitter for visibility. Brackets above each panel show paired-sample t-tests for the three consecutive comparisons, with significance markers (*** $p < 0.001$, ** $p < 0.01$, * $p < 0.05$) based on Benjamini–Hochberg false-discovery-rate correction.*

## Section S2. Diagnosis-Level Prediction Diagnostics

Beyond the syndrome-level analysis reported in the main text, we examined the joint structure of clinical and predicted variables at the individual-patient level. Figure S2 shows the Spearman correlation matrix among WAB-AQ, correct response rate, and the predicted perturbation parameters; predicted modification percentage correlates most strongly with the correct response rate ($r = -0.38$), consistent with perturbation intensity rather than layer carrying the behaviorally salient signal. Figure S3 shows the distribution of predicted layer indices across the 278 CSTAR patients, both overall and stratified by diagnosis; the predicted-layer distribution is broadly consistent across aphasia subtypes, indicating that the layer prediction does not trivially encode diagnostic category.

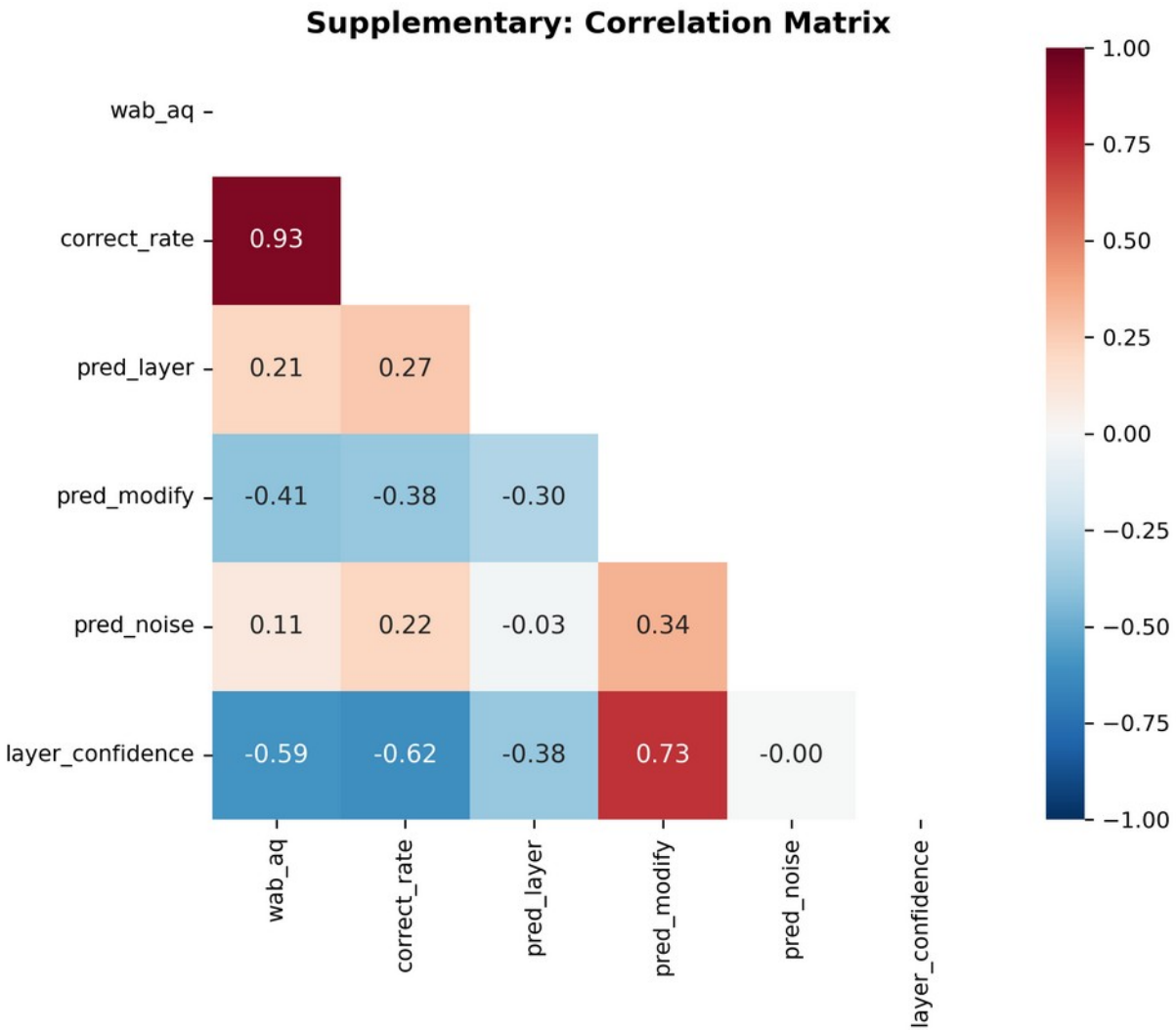

***Figure S2.*** *Correlation Matrix of Clinical and Predicted Variables. Heatmap showing Spearman correlations among WAB-AQ, correct response rate, predicted layer, predicted modification percentage, predicted noise sigma, and layer prediction confidence. The matrix permits inspection of the joint structure across clinical (WAB-AQ, correct response rate) and predicted (perturbation parameters) variables; main-text analyses focus on syndrome-level discrimination rather than continuous severity correlations. Predicted modification percentage correlates most strongly with correct rate ($r = -0.38$).*

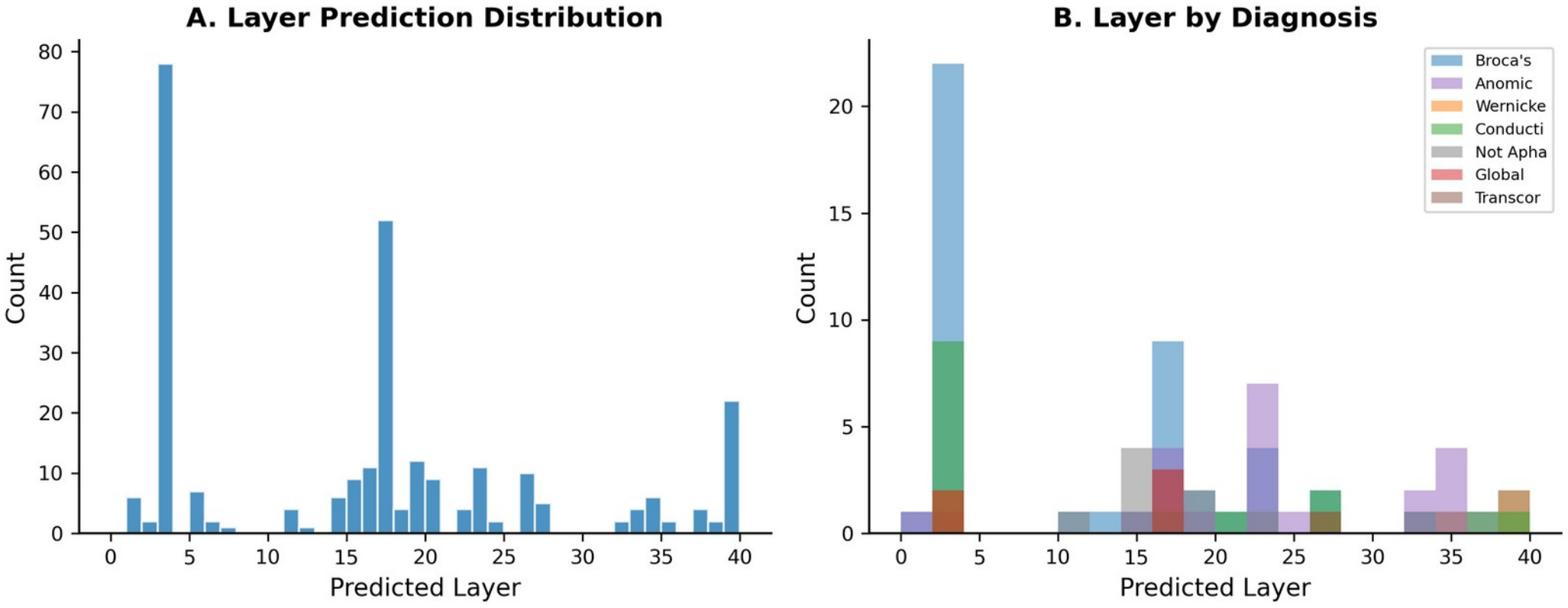

***Figure S3.*** *Layer Prediction Distribution. (A) Overall histogram of predicted layer indices across 278 patients from the CSTAR dataset. (B) Stacked histogram by diagnosis category, demonstrating consistent layer prediction patterns across aphasia subtypes.*

## Section S3. Syndrome Signature Analysis

Supplementary figures and the full pairwise-contrast table for the syndrome signature analysis described in the main text. Sample, group sizes, and statistical methods are identical to those reported in the Results.

The syndrome-level differences reported in the main text are visualized per parameter in Figures S4–S6. Predicted layer index (Figure S4) is earliest for the most severe syndromes — Global (mean 11.2) and Broca's (12.0) — though the ordering is not strictly monotonic across milder groups (Kruskal–Wallis $\eta^2 = 0.130$). Predicted noise intensity (Figure S5) does not differ significantly across syndrome groups ($\eta^2 = 0.00$, n.s.). The severity gradient (Figure S6) shows that modification percentage trends downward against WAB-AQ across syndromes, whereas noise intensity shows no consistent trend. Full pairwise contrasts are given in Table S3.

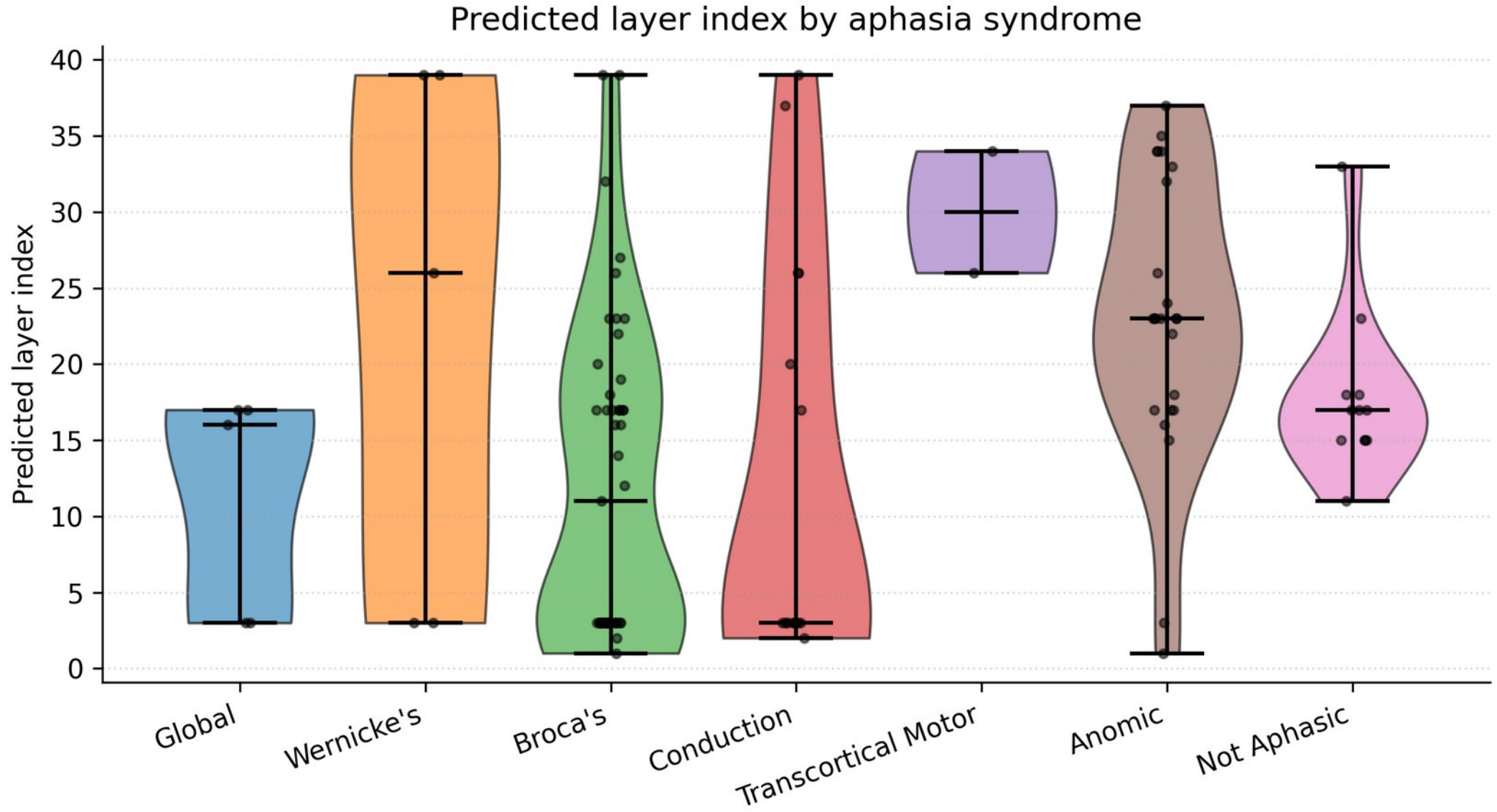

***Figure S4.*** *Predicted layer index by aphasia syndrome. Violin plots show the distribution of inverse-model-predicted transformer layer for each syndrome group. The most severe syndromes map to the earliest layers (Global 11.2, Broca's 12.0); the ordering is not strictly monotonic across milder groups. Kruskal–Wallis $H = 18.25$, $p = 0.003$, $\eta^2 = 0.130$.*

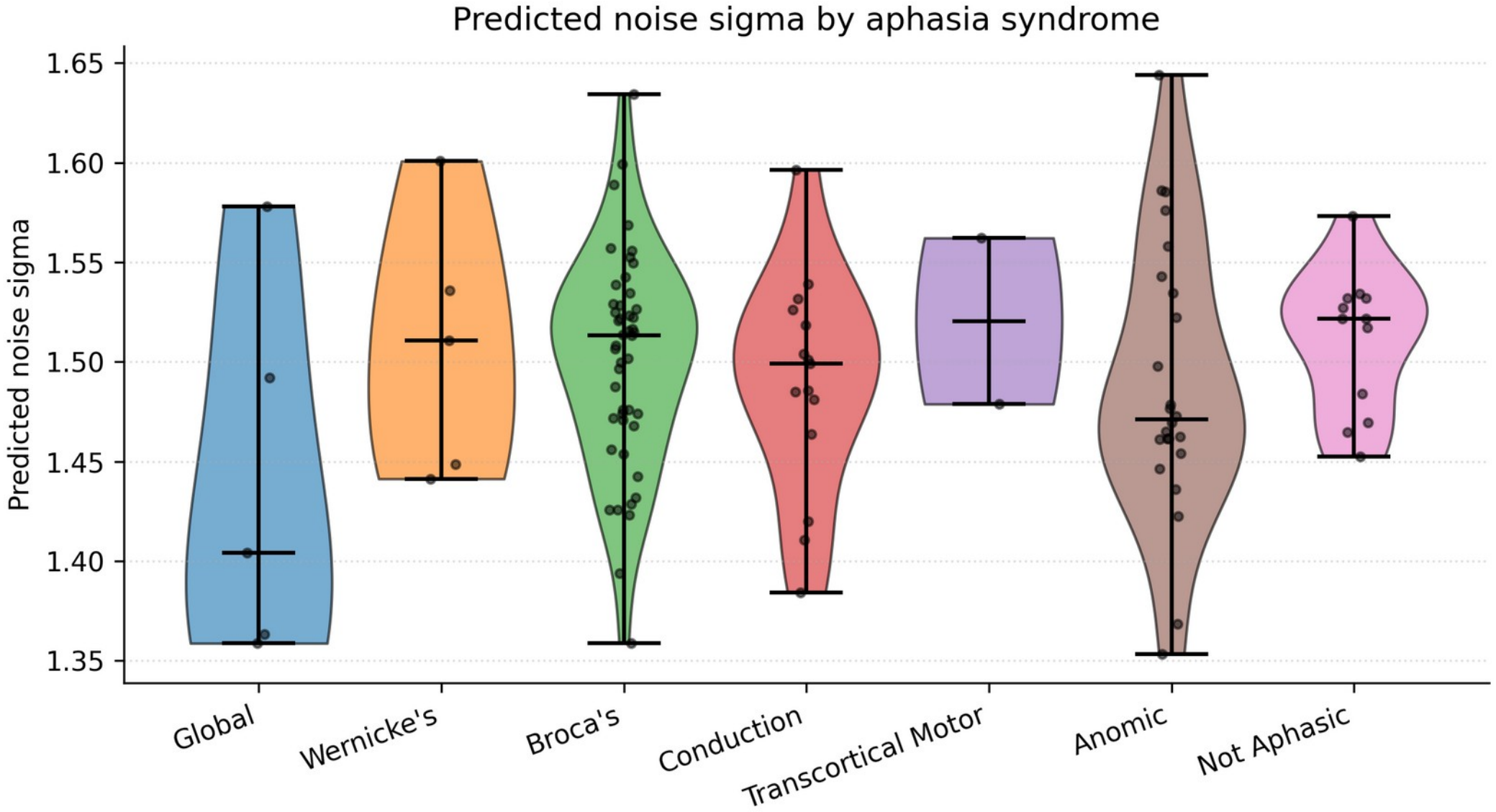

***Figure S5.*** *Predicted noise sigma by aphasia syndrome. Violin plots show the distribution of predicted noise intensity for each syndrome group. Kruskal–Wallis H = 4.63, p = 0.46, η² = 0.00 (not significant).*

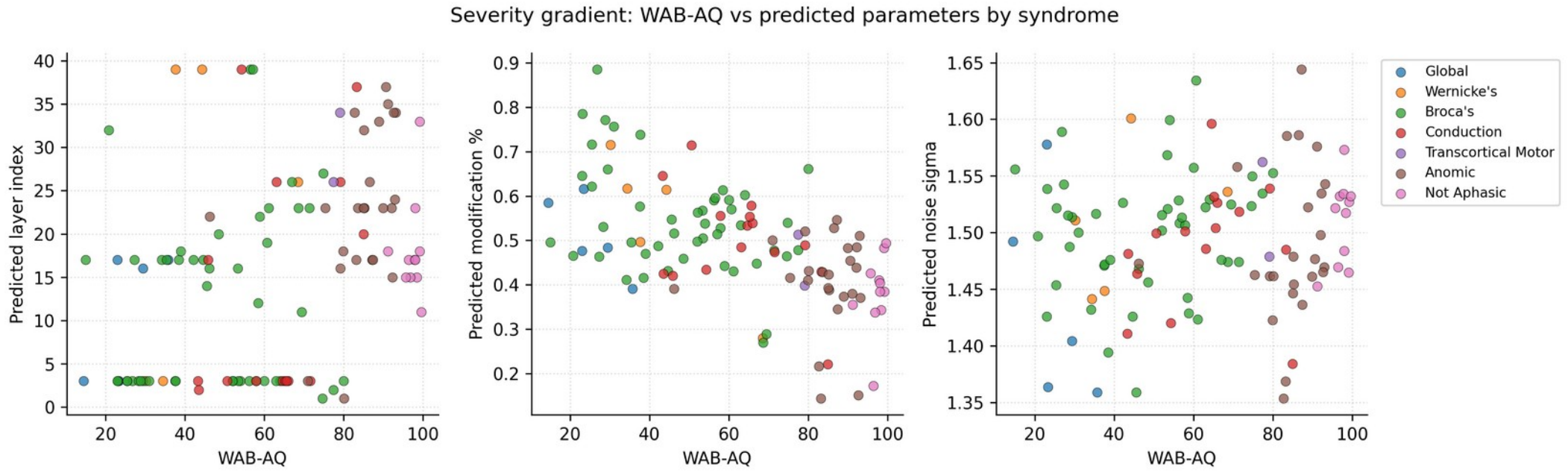

***Figure S6.*** *Severity gradient: WAB-AQ versus the three predicted perturbation parameters, color-coded by aphasia syndrome (n = 110 with diagnostic labels). Each panel shows individual patients; the downward trend in modification % against WAB-AQ confirms the negative severity correlation, whereas noise sigma shows no consistent trend.*

**Table S3.** *Significant pairwise Mann–Whitney U contrasts (Benjamini–Hochberg corrected p < 0.05) for syndrome-level differences in predicted perturbation parameters. Cliff's δ in [−1, 1] is the non-parametric effect size; values near ±1 indicate near-perfect group separation.*

| Parameter | Group A | Group B | $n_a$ | $n^b$ | p (BH) | Cliff's δ |
|---|---|---|---|---|---|---|
| Modification % | Anomic | Broca's | 24 | 47 | < 0.0001 | −0.67 |
| Modification % | Anomic | Conduction | 24 | 15 | 0.020 | −0.54 |
| Modification % | Broca's | Not Aphasic | 47 | 12 | < 0.001 | +0.81 |
| Modification % | Conduction | Not Aphasic | 15 | 12 | 0.011 | +0.70 |
| Layer index | Anomic | Broca's | 24 | 47 | 0.002 | +0.55 |

## Section S4. Sensitivity and Robustness Analysis

Additional visualizations for the sensitivity analysis described in the main text and Table 5 are provided. The subsample power curves (Figure S7) show the proportion of stratified bootstrap replicates yielding a significant omnibus test as a function of subsample fraction, for each parameter and each grouping condition; modification percentage reaches near-ceiling detection even at the smallest subsamples, and layer index requires larger subsamples to reach the 80% detection target; predicted noise sigma, being non-significant across conditions, does not reach the detection target. The bootstrap η² distribution (Figure S8) is shown for the primary condition (C1) as an illustrative example; analogous distributions for C2–C5 follow the same pattern and are available in the project data release.

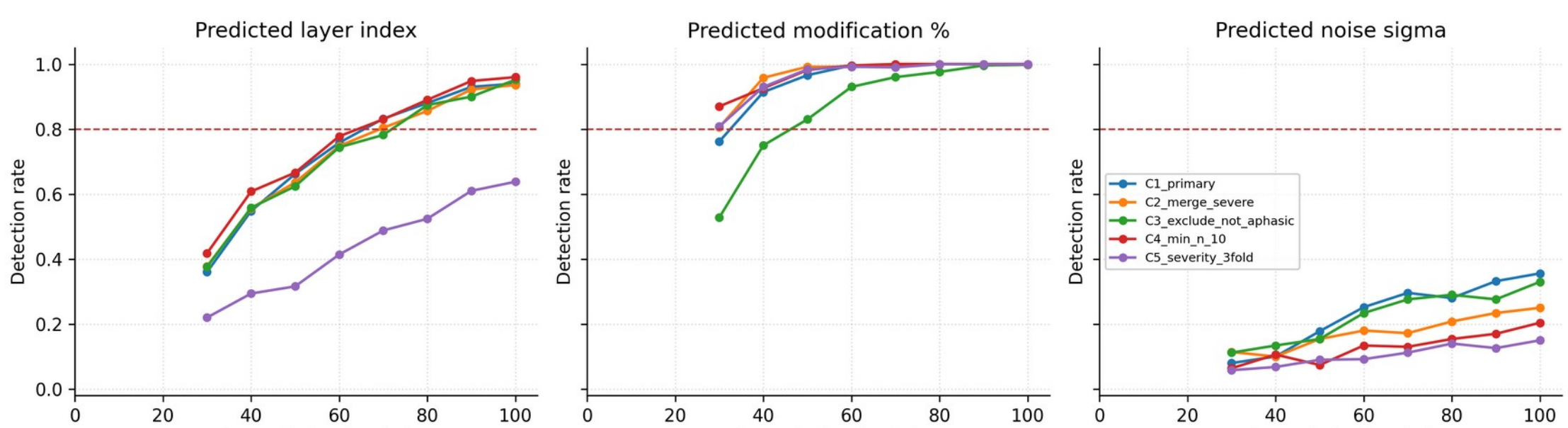

***Figure S7.*** *Distribution-free subsample power curves across sensitivity conditions. For each condition and each parameter, we drew stratified bootstrap subsamples at fractions 30–100% of each group's original size (2,000 replicates per fraction) and computed the proportion of replicates yielding omnibus p < 0.05. The red dashed line marks the conventional 80% detection target. Modification percentage (center panel) reaches near-ceiling detection at even 30% subsample, and layer index reaches 80% detection at 40–50% subsample fraction in most conditions; predicted noise sigma does not reach the 80% target (non-significant across conditions). The C3 condition (exclude Not Aphasic) is the slowest to saturate, consistent with its reduced effect sizes.*

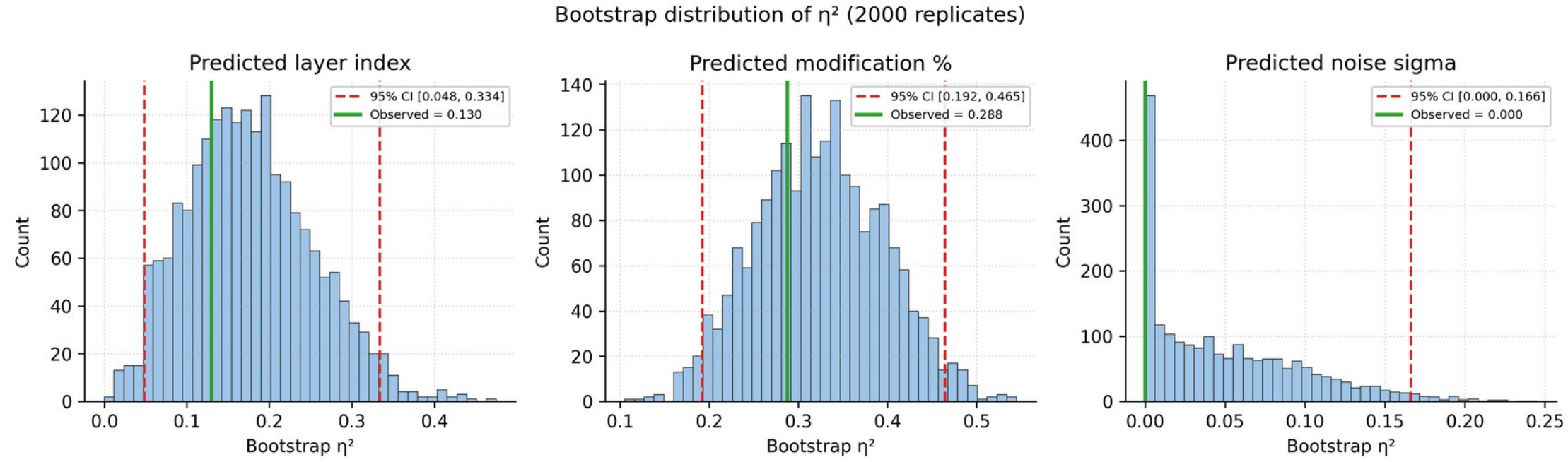

***Figure S8.*** *Bootstrap η² distributions for the primary sensitivity condition (C1, six syndrome groups, n = 108). Histograms show 2,000 stratified bootstrap replicates per parameter; red dashed lines mark the 2.5% and 97.5% percentiles (95% confidence interval), and the green solid line marks the observed η². The modification-percentage confidence interval lies entirely above the medium-effect threshold (η² = 0.06); the layer-index interval lies above the small-effect threshold (η² = 0.01); the noise interval includes zero, consistent with a non-significant effect.*

**Table S4.** *Full sensitivity analysis results, split into three per-parameter sub-tables (Modification %, Layer index, Noise sigma). For each of the five grouping conditions, the table reports the Kruskal–Wallis H statistic, p-value, 10,000-iteration permutation p-value, eta-squared effect size, bootstrap 95% confidence interval, number of significant pairwise contrasts (BH-corrected), minimum detectable effect at 80% power, and achieved power at the observed effect size.*

**Table S4a:** ***Modification %***

| Condition | k | n | H | p (KW) | Perm p | η² | 95% CI | Sig | MDE | Power |
|---|---|---|---|---|---|---|---|---|---|---|
| C1 primary | 6 | 108 | 34.34 | < 0.001 | < 0.001 | 0.288 | [0.192, 0.465] | 4/15 | 0.119 | >0.99 |
| C2 merge Wer/Glo | 5 | 108 | 34.03 | < 0.001 | < 0.001 | 0.292 | [0.186, 0.450] | 6/10 | 0.111 | >0.99 |
| C3 excl. Not Aph. | 5 | 96 | 22.38 | < 0.001 | < 0.001 | 0.202 | [0.111, 0.390] | 2/10 | 0.124 | 0.958 |
| C4 min n = 10 | 4 | 98 | 33.62 | < 0.001 | < 0.001 | 0.326 | [0.205, 0.492] | 4/6 | 0.111 | >0.99 |
| C5 severity 3-fold | 3 | 108 | 27.66 | < 0.001 | < 0.001 | 0.244 | [0.132, 0.394] | 3/3 | 0.089 | >0.99 |

**Table S4b:** ***Layer index***

| Condition | k | n | H | p (KW) | Perm p | η² | 95% CI | Sig | MDE | Power |
|---|---|---|---|---|---|---|---|---|---|---|
| C1 primary | 6 | 108 | 18.25 | 0.003 | < 0.001 | 0.130 | [0.048, 0.334] | 1/15 | 0.119 | 0.839 |

| Condition | k | n | H | p (KW) | Perm p | $\eta^2$ | 95% CI | Sig | MDE | Power |
|---|---|---|---|---|---|---|---|---|---|---|
| C2 merge Wer/Glo | 5 | 108 | 16.37 | 0.003 | < 0.001 | 0.120 | [0.037, 0.301] | 1/10 | 0.111 | 0.836 |
| C3 excl. Not Aph. | 5 | 96 | 16.09 | 0.003 | 0.002 | 0.133 | [0.040, 0.347] | 1/10 | 0.124 | 0.829 |
| C4 min n = 10 | 4 | 98 | 16.76 | < 0.001 | < 0.001 | 0.146 | [0.040, 0.340] | 1/6 | 0.111 | 0.904 |
| C5 severity 3-fold | 3 | 108 | 7.48 | 0.024 | 0.025 | 0.052 | [0.000, 0.199] | 1/3 | 0.089 | 0.556 |

**Table S4c:** ***Noise sigma***

| Condition | k | n | H | p (KW) | Perm p | $\eta^2$ | 95% CI | Sig | MDE | Power |
|---|---|---|---|---|---|---|---|---|---|---|
| C1 primary | 6 | 108 | 4.63 | 0.46 | 0.478 | 0.000 | [0.000, 0.166] | 0/15 | 0.119 | — |
| C2 merge Wer/Glo | 5 | 108 | 2.99 | 0.56 | 0.575 | 0.000 | [0.000, 0.128] | 0/10 | 0.111 | — |
| C3 excl. Not Aph. | 5 | 96 | 3.68 | 0.45 | 0.460 | 0.000 | [0.000, 0.161] | 0/10 | 0.124 | — |
| C4 min n = 10 | 4 | 98 | 2.27 | 0.52 | 0.527 | 0.000 | [0.000, 0.117] | 0/6 | 0.111 | — |
| C5 severity 3-fold | 3 | 108 | 1.77 | 0.41 | 0.419 | 0.000 | [0.000, 0.084] | 0/3 | 0.089 | — |

**Note.** *Condition abbreviations as in Table 5. MDE = minimum detectable $\eta^2$ at 80% power and $\alpha = 0.05$, computed from the non-central $\chi^2$ approximation. Sig = number of pairwise Mann–Whitney U contrasts significant at Benjamini–Hochberg-corrected $p < 0.05$ out of the total number of possible pairs. Permutation p-values reported as < 0.001 reflect the 10,000-iteration resolution limit (add-one correction applied).*

## Section S5. Multi-Seed Robustness Analysis

This section reports the per-layer accuracy structure of the inverse model averaged across 10 independently trained instances. The multi-seed evaluation directly addresses whether the conclusions in the main paper depend on stochastic factors of a single training run. Across 10 independently trained inverse models, the within-5 layer accuracy on the in-distribution test set is 77.6% ± 1.1% and the modification $R^2$ is 0.570 ± 0.025 (Table 3 of the main paper); on the full cross-seed held-out grid (n = 4,840 configurations), single-model layer top-1 accuracy is 16.3% ± 0.3% and within-5 is 78.1% ± 0.6% (Table 3b of the main paper). The cross-seed standard deviations are tighter than the in-distribution test variability, indicating the inverse mapping generalizes to new stochastic realizations of the training data rather than overfitting to any single training seed.

Below the aggregate metrics, the per-layer accuracy structure (Figure S9) is itself informative: top-1 accuracy is markedly elevated for the embedding-adjacent layers (layers 0–2) and for select late-stack layers (layers 21–22, 36–39), while mid-stack layers (layers 3–18) show accuracy near or modestly above the 2.5% random baseline. The pattern is consistent across all 10 training seeds (error bars in Figure S9), with the largest error bars concentrated on the early and late layers where mean accuracy is highest. Under the asymmetric U-shaped recoverability framing introduced in the Discussion of the main paper, this is consistent with vision–language interface and output projection layers carrying more distinctive behavioral signatures than computational mid-stack layers.

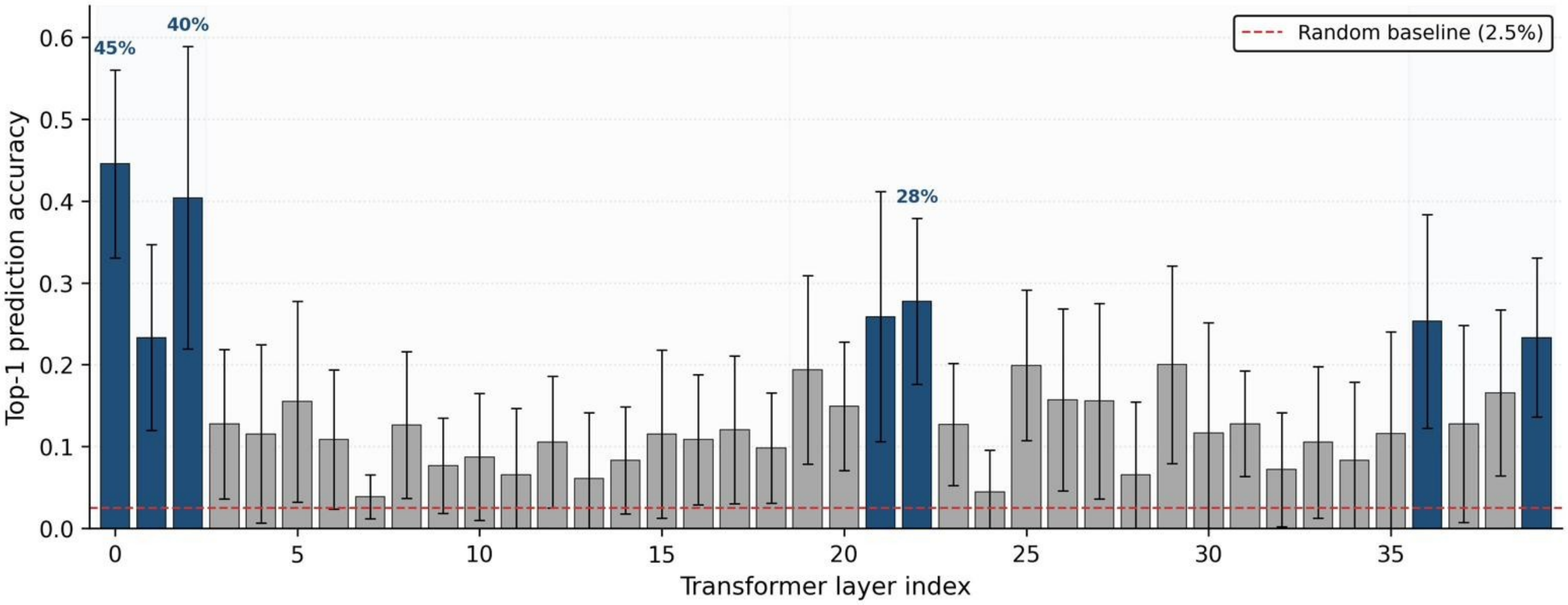

***Figure S9.*** *Per-layer top-1 prediction accuracy across 10 independently trained inverse models. Bars show mean accuracy for each of the 40 transformer layers; error bars represent standard deviation across the 10 training seeds. The horizontal red dashed line marks the 2.5% chance baseline (uniform random prediction over 40 layers). Embedding-adjacent layers (0–2) and select late-stack layers (22, 36–39) are highlighted in dark blue; the top three layers achieve mean accuracies of 45% (layer 0), 40% (layer 2), and 28% (layer 22). Mid-stack layers (3–18) cluster near or modestly above the chance baseline, consistent with functional redundancy in the computational middle of the network.*

Aggregate per-region binning across the 40 layers is summarized in Table S5. Layer regions are defined as Early (layers 0–2), Mid-early (layers 3–18), Mid-late (layers 19–35), and Late (layers 36–39).

***Table S5.*** *Mean top-1 layer-prediction accuracy by layer region, aggregated across 10 independently trained inverse models. Each region's mean is computed first by averaging within-region per-layer accuracies for each seed, then taking the mean across seeds. Number of layers in each region is shown for reference.*

| Layer region | Layer indices | N layers | Mean top-1 accuracy | Chance |
|---|---|---|---|---|
| **Early** | 0–2 | 3 | 36.1% | 2.5% |
| **Mid-early** | 3–18 | 16 | 10.0% | 2.5% |
| **Mid-late** | 19–35 | 17 | 14.4% | 2.5% |
| **Late** | 36–39 | 4 | 19.5% | 2.5% |

**Note.** *Per-region means provide a coarse summary of the asymmetric U-shaped recoverability pattern visible in Figure S9. Per-seed individual metrics for all 10 training runs are included in the data release accompanying this manuscript.*

## Section S6. Counterfactual Baseline Ablation

This section reports an ablation baseline for the counterfactual reproduction fidelity. Three arms were evaluated on the identical 4,840 held-out cross-seed configurations, paired by configuration identifier: the unaltered pipeline (A), an arm that randomizes the predicted layer while retaining predicted intensity (C), and an arm that replaces the predicted per-case intensity with the dataset-wide mean while retaining the predicted layer (D). The arms differ only in the substituted quantity; no inverse model was retrained. Cosine similarity and error correlation are bounded at 1.0 and strongly left-skewed, so central tendency is reported as median with interquartile range. This cross-seed replication yields 79.9% high-fidelity for the unaltered arm, reproducing the main-text validation result (81.4%, Table 2) within seed-level variation.

### S6.1 Fidelity by arm

| Arm | Layer | Intensity | HIGH | MODERATE | LOW | Cosine median [IQR] |
|---|---|---|---|---|---|---|
| A | predicted | predicted | 79.9% | 8.4% | 11.7% | 0.998 [0.951, 1.000] |
| C | randomized | predicted | 62.8% | 6.5% | 30.6% | 0.992 [0.594, 1.000] |
| D | predicted | marginal | 42.4% | 8.0% | 49.6% | 0.781 [0.135, 0.996] |

*Table S6.1. Reproduction fidelity across the three baseline arms on the identical 4,840 held-out cross-seed configurations. HIGH/MODERATE/LOW are percentages of cases. Paired Wilcoxon signed-rank tests: A versus C, Δcosine = 0.142, $p = 2.2\times10^{-166}$; A versus D, Δcosine = 0.316, $p < 10^{-300}$; C versus D, Δcosine = 0.174, $p = 9.9\times10^{-118}$.*

### S6.2 Stratified analysis

Stratifying the layer-randomization arm (C) by the depth of the originally targeted layer shows the largest fidelity loss for the shallowest layers, consistent with greater functional redundancy among deeper layers. Stratifying the marginal-intensity arm (D) by true modification strength shows the largest loss at the lowest strengths, where the dataset-average intensity most overshoots the true perturbation.

| True layer bucket | n | Arm A cosine | Arm C cosine | Cosine drop |
|---|---|---|---|---|
| 0-9 (shallow) | 1210 | 0.889 | 0.705 | 0.184 |
| 10-19 (mid) | 1210 | 0.885 | 0.761 | 0.123 |
| 20-29 (deep) | 1210 | 0.943 | 0.831 | 0.112 |
| 30-39 (deepest) | 1210 | 0.965 | 0.818 | 0.147 |

*Table S6.2. Mean cosine similarity for the full arm (A) and the random-layer arm (C), stratified by the depth of the originally targeted layer.*

| True modification | n | Arm A cosine | Arm D cosine | Cosine drop |
|---|---|---|---|---|
| 0.0-0.2 | 1320 | 0.971 | 0.483 | 0.488 |
| 0.3-0.5 | 1320 | 0.908 | 0.740 | 0.168 |
| 0.6-0.8 | 1320 | 0.896 | 0.629 | 0.267 |
| 0.9-1.0 | 880 | 0.900 | 0.547 | 0.353 |

*Table S6.3. Mean cosine similarity for the full arm (A) and the marginal-intensity arm (D), stratified by true modification strength.*

### S6.3 Fidelity distributions

Table S6.1 reports central tendency, but the per-case distributions in Figure S10 reveal that the two ablations degrade reproduction through different mechanisms. Randomizing the predicted layer (arm C) leaves the upper portion of the distribution largely intact and instead lengthens the lower tail: most cases remain faithfully reproduced, and the loss in mean fidelity is driven by a subset of cases for which the substituted layer falls outside the functionally equivalent range. Replacing the predicted intensity with the dataset-wide mean (arm D) instead shifts the entire distribution downward, degrading reproduction across the full range of cases rather than a subset. This distributional contrast explains why the intensity ablation is the more damaging of the two, and it is consistent with the interpretation that layer identity is recoverable only up to an equivalence class, whereas intensity must be recovered on a per-case basis.

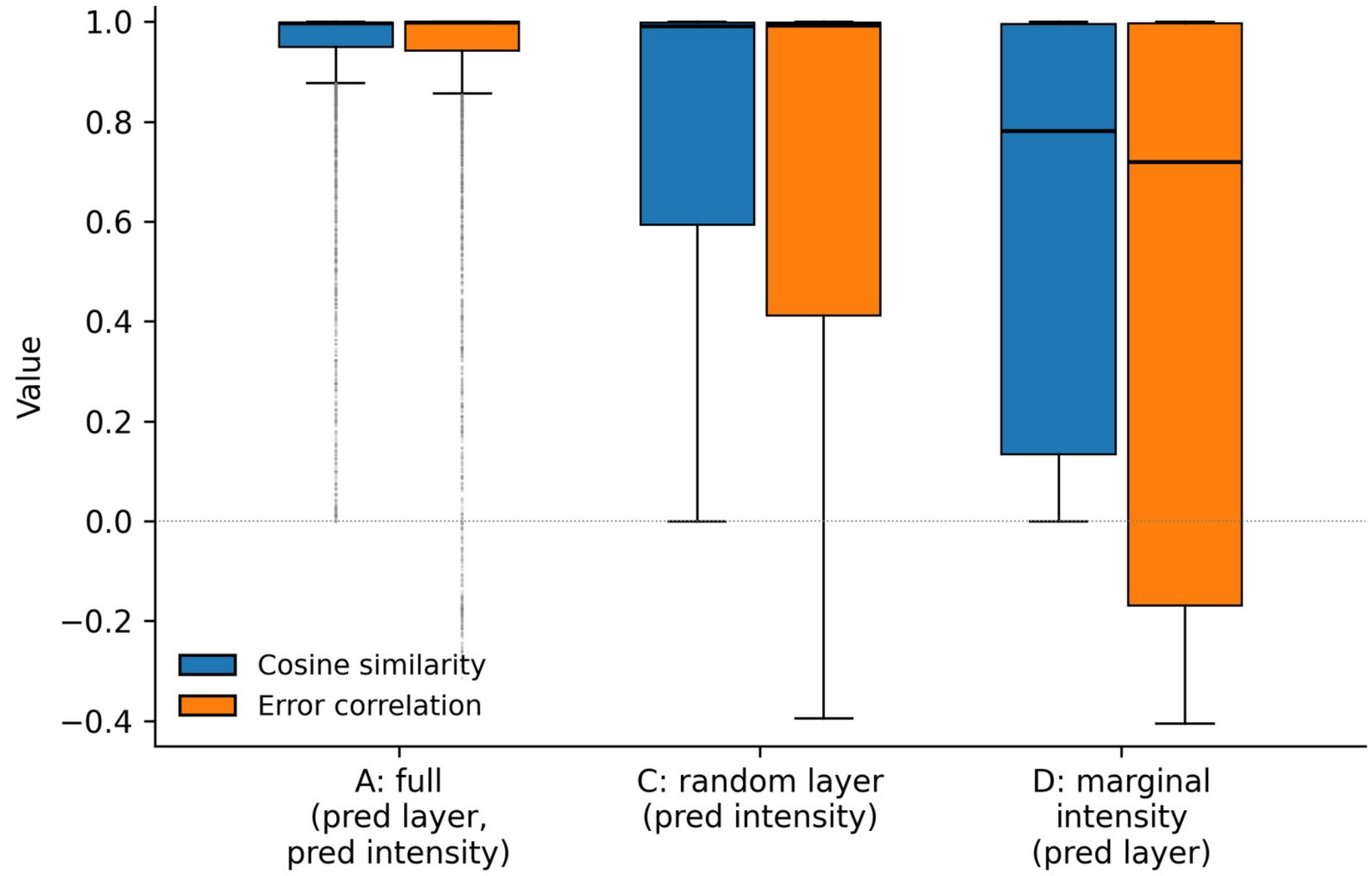

***Figure S10.*** *Per-case cosine similarity (blue) and error correlation (orange) by arm (box = interquartile range, line = median, whiskers = 1.5 IQR, points = outliers). Randomizing the layer (C) lengthens the lower tail; marginalizing intensity (D) shifts the whole distribution downward.*